\documentclass[10pt,twocolumn,table,xcdraw,letterpaper]{article}

\usepackage[pagenumbers]{cvpr} 

\definecolor{cvprblue}{rgb}{0.21,0.49,0.74}
\usepackage[accsupp]{axessibility}  
\usepackage[pagebackref,breaklinks,colorlinks,allcolors=cvprblue]{hyperref}
\usepackage{booktabs}
\usepackage{color}
\usepackage{multirow}
\usepackage{array}
\usepackage{algorithm}
\usepackage[noend]{algpseudocode}
\usepackage{bbm}
\usepackage{adjustbox}
\usepackage{makecell}
\usepackage{tabularx}
\usepackage[normalem]{ulem}
\newcommand{\methodname}{ASCED}
\definecolor{mygray}{gray}{.9}
\definecolor{myred}{rgb}{0.81, 0.0, 0.0}
\definecolor{mygreen}{rgb}{0.0, 0.51, 0.0}
\definecolor{myyellow}{rgb}{0.71, 0.55, 0.0}
\definecolor{myblue}{rgb}{0.0, 0.71, 0.71}
\definecolor{color2}{rgb}{0.55, 0.71, 0.0}
\definecolor{color1}{rgb}{0.98, 0.81, 0.69}
\definecolor{color3}{rgb}{1.0, 0.6, 0.4}
\definecolor{color4}{rgb}{0.29, 0.59, 0.82}

\def\paperID{2601} 
\def\confName{CVPR}
\def\confYear{2025}

\title{Temporal Score Analysis for Understanding and Correcting Diffusion Artifacts}

\author{Yu Cao \hspace{1.5em}
 Zengqun Zhao \hspace{1.5em}
 Ioannis Patras \hspace{1.5em}
 Shaogang Gong\\
Queen Mary University of London\\
{\tt\small \{yu.cao, zengqun.zhao, i.patras, s.gong\}@qmul.ac.uk}
}

\begin{document}

\twocolumn[{
\renewcommand\twocolumn[1][]{#1}
\maketitle
\begin{center}
    \captionsetup{type=figure}
    \vspace{-1 em}
    \includegraphics[width=1\linewidth]{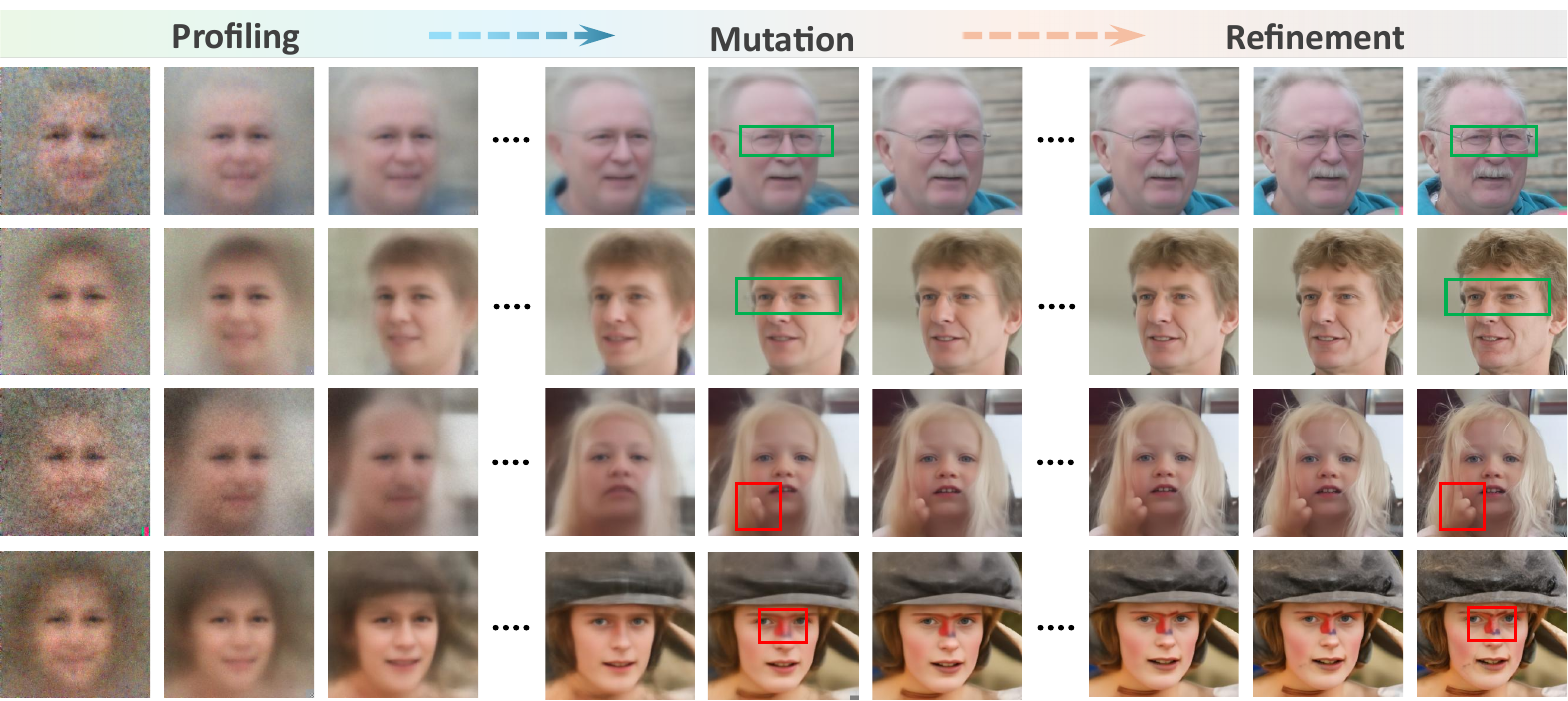}
    \vspace{-1.5 em}
    \captionof{figure}{\textbf{Why do diffusion models generate artifacts?} We discover that a diffusion generative process necessarily undergoes three phases, we call them: (2) ``Profiling" which recovers holistic mean templates, (2) ``Mutation" which introduces local divergence, and (3) ``Refinement" which rationalizes pixel-wise generation in spatial context. Four visual examples are shown: The first two rows are two examples of rational local mutations (in green boxes) either naturally integrated (Row 1) or reasonably eliminated (Row 2). The bottom two rows show two failure cases when mutations were trapped unreasonably (in red boxes), resisting refinement and resulting in artifacts. Phases are visualized in equal intervals for clarity; please zoom in for more details.
    }
    \label{fig:illustration}
\end{center}
}]

\begin{abstract}
Visual artifacts remain a persistent challenge in diffusion models, even with training on massive datasets. Current solutions primarily rely on supervised detectors, yet lack understanding of why these artifacts occur in the first place. 
In our analysis, we identify three distinct phases in the diffusion generative process: Profiling, Mutation, and Refinement. Artifacts typically emerge during the Mutation phase, where certain regions exhibit anomalous score dynamics over time, causing abrupt disruptions in the normal evolution pattern. 
This temporal nature explains why existing methods focusing only on spatial uncertainty of the final output fail at effective artifact localization.
Based on these insights, we propose \methodname~(Abnormal Score Correction for Enhancing Diffusion), that detects artifacts by monitoring abnormal score dynamics during the diffusion process, with a trajectory-aware on-the-fly mitigation strategy that appropriate generation of noise in the detected areas. 
Unlike most existing methods that apply post hoc corrections, \eg, by applying a noising-denoising scheme after generation, our mitigation strategy operates seamlessly within the existing diffusion process. 
Extensive experiments demonstrate that our proposed approach effectively reduces artifacts across diverse domains, matching or surpassing existing supervised methods without additional training.
Project page: \href{https://YuCao16.github.io/ASCED/}{YuCao16.github.io/ASCED}.
\end{abstract}

\vspace{-2 em}
\section{Introduction}

\begin{figure*}[tb]
\centering
\vspace{-1 em}
\includegraphics[width=0.88\linewidth]{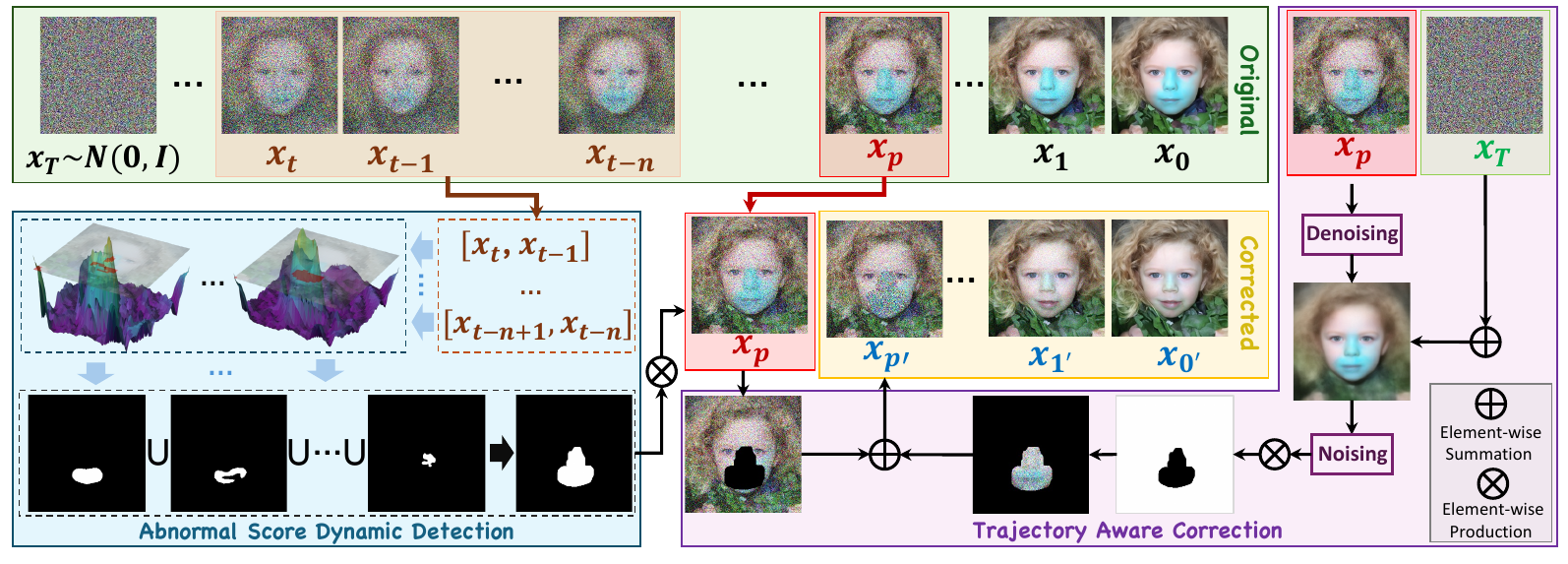}
\vspace{-1 em}
\caption{\textbf{Diagram of our framework}. Denoising and Noising are using \cref{eq:pred_xstart} and \cref{eq:forward_sde}, respectively.}
\vspace{-1.5 em}
\label{fig:framework}
\end{figure*}

Diffusion models have emerged as powerful foundation models in computer vision \cite{bommasani2021opportunities}, achieving remarkable success in image generation \cite{higgins2016beta, karras2019style, ho2020denoising,saharia2022photorealistic, cao2024few}, image inpainting \cite{song2020score, meng2021sdedit, lugmayr2022repaint}, and text-to-image task \cite{saharia2022photorealistic, rombach2022high, ruiz2023dreambooth}. 
However, even trained on large-scale datasets, diffusion generative images still exhibit two significant flaws: visual artifacts and hallucinations \cite{zhang2023perceptual, zhang2023siren}. 
Visual artifacts appear as local irregularities in texture or structure, while hallucinations involve semantically incoherent content, \eg, extra limbs or misplaced objects. In this work, we focus on addressing diffusion artifacts, which present a fundamental challenge to achieving reliable and high-quality image generation.

Existing methods primarily treat visual artifact detection as a classification problem, \ie, identifying problematic generations for filtering or reconstruction. These methods typically rely on a specialized classifier, either trained on manually annotated artifact datasets \cite{zhang2023perceptual} or leveraging a pre-trained Large Multi-Modal Model (LMM) \cite{liu2024improved}.  
However, such post hoc interventions fail to address a fundamental problem: \textbf{Why and when do artifacts emerge in diffusion models?} To bridge this gap, we begin by examining the diffusion generation process itself.

We discover that while diffusion process is guided by the same fundamental equation across time (\ie diffusion steps), in practice, the model exhibits different behavior that can be roughly categorized in three different temporal phases that we name Profiling, Mutation, and Refinement. In the ``Profiling'' phase, the model sketches the basic semantic global layout; in the ``Mutation'' phase it explores potential local pixel-wise variations to create local structure; in the ``Refinement'' phase it attempts to resolve these local pixel-wise variations into coherent visual details in context (see \cref{fig:illustration} for visual examples and \cref{sec:identification} for a detailed analysis).
This understanding of the generation process reveals that while visual artifacts may appear randomly, they follow systematic and identifiable temporal patterns during image formation.
Recent uncertainty-based approaches try to identify artifacts by converting diffusion models into Bayesian networks and employ techniques such as Last Layer Laplace Approximation \cite{daxberger2021laplace} to generate pixel-level variance matrices \cite{kou2023bayesdiff}. However, these uncertainty quantification analyses only capture spatial variations in the final output, \ie, $Var(x_0)$, missing crucial temporal dynamics during the generation process.
Our study shows that diffusion artifacts emerge when certain image regions exhibit abnormal evolution patterns \textit{over time}, primarily during the Mutation phase of the generation process. Specifically, their pixel values stop updating properly while the surrounding areas continue to evolve. This phenomenon, which we formally define as ``score traps" in \cref{sec:identification}, explains why examining only the final output is insufficient and misplaced for effective diffusion artifact detection.

Building on these insights, we propose \methodname: Abnormal Score Correction for Enhancing Diffusion, as shown in \cref{fig:framework}. At the heart of the method is the estimation of a score, in this context the direction and magnitude of pixel-wise evolution at each diffusion generative step, and a scheme that analyses its temporal dynamics and detects abnormalities. 
We show both theoretically and experimentally that these abnormalities strongly correlate with artifact formation, making early detection possible at a stage where intervention is still feasible, before artifacts become irreversibly embedded in the generation process.
We leverage this early detection capability by implementing a novel trajectory-aware correction mechanism that disrupts the evolution of artifact regions while preserving overall generation diversity. Importantly, \methodname~operates in a fully unsupervised manner without requiring manual annotations or domain-specific training, making it readily applicable across various domains, , particularly valuable when training data may be limited or protected.

Our contributions are: \textbf{(1)} We provide new insights into the formation of visual artifacts in the diffusion generative process, advancing the understanding of diffusion model internal mechanisms. \textbf{(2)} We introduce a novel method that detects potential artifact regions by monitoring abnormal score dynamics temporally, without any manually labeled training required. \textbf{(3)} We further develop a on-the-fly trajectory-aware correction mechanism that effectively mitigates artifacts while preserving image diversity. 

\section{Related Works}

\noindent\textbf{Visual Artifact Detection} 
initially targeted super-resolution artifacts, where upsampling operations are the main source \cite{zhang2019detecting}.
These methods analyze either spatial domain characteristics to capture texture differences between real and generated images \cite{yu2019attributing, liu2020global}, or frequency domain patterns to study artifact characteristics in high-frequency components \cite{durall2020watch, dzanic2020fourier}.
More recent work has shifted focus to detecting artifacts in general image generation, developing specialized classifiers trained on manually annotated datasets \cite{zhang2023perceptual} or utilizing pre-trained large vision models \cite{liu2024improved}. However, these supervised approaches require extensive manual labeling and may not generalize well across different domains.
A parallel direction explores uncertainty quantification methods to understand visual artifacts. While various approaches including variational inference \cite{blundell2015weight, khan2018fast}, Laplace approximation \cite{mackay1992bayesian, ritter2018scalable}, and Markov Chain Monte Carlo \cite{welling2011bayesian, zhang2019cyclical} have been developed, their application to diffusion models remains limited \cite{kou2023bayesdiff}. BayesDiff \cite{kou2023bayesdiff} pioneers pixel-level uncertainty quantification in diffusion models using Last-layer Laplace approximation \cite{daxberger2021laplace}, yet the connection between spatial uncertainty and visual artifacts remains unclear.

\noindent\textbf{Generation Quality Enhancement}
Various approaches have been proposed to enhance generation quality. The truncation trick in BigGAN \cite{brock2018large} demonstrated that restricting the sampling space can significantly improve generation fidelity, suggesting similar principles might apply to diffusion models. For diffusion-based generation, classifier guidance \cite{dhariwal2021diffusion} has been introduced to steer the generation process.
SARGD \cite{zheng2024self} extends this idea by utilizing a pre-trained artifact detector to guide the generation towards artifact-free regions.
Latent diffusion models \cite{rombach2022high} take a different approach by applying diffusion in a learned latent space, demonstrating that controlled evolution in a constrained space can lead to higher-quality output. 

\noindent\textbf{Diffusion Model for Representation Learning}
has evolved toward latent space disentanglement and controllable editing. The former aims to uncover interpretable factors in the generative process. Recent studies \cite{zhang2022unsupervised, preechakul2022diffusion, yue2024exploring} observed stage-wise attribute emergence during generation, but focus differently than our analysis of artifact formation mechanisms.
Controllable editing techniques \cite{wallace2023edict, zhang2023inversion, meng2021sdedit, ruiz2023dreambooth} can be applied for artifact removal, yet require per-sample manipulation and address symptoms rather than causes. Our approach instead corrects abnormal score dynamics during the generation process itself.

\section{Methodology}

Our approach consists two steps: Detection (\cref{sec:identification}) and Correction (\cref{sec:correction}). Specifically, we localize artifact pixel regions by identifying abnormal score dynamics during the diffusion inference process (Detection) and develop a novel artifact correction algorithm without delaying inference (Correction). We give a theoretical analysis of our key concepts on score trap and temporal weighting (\cref{sec:theoretical}).

\subsection{Preliminaries}

\paragraph{Diffusion Model} Let $x_0 \in \mathbb{R}^{c \times h \times w}$ be an image. The forward process of a diffusion model gradually diffuses the data distribution $q(x)$ towards $q_t(x_t)$, $\forall t \in [0, T]$, with $q_T(x_T) = \mathcal{N}(\boldsymbol{0}, \boldsymbol{I})$ as a trivial Gaussian distribution. From a score viewpoint, it can be described by the Stochastic Differential Equation (SDE):
\vspace{-0.5 em}
\begin{equation}
    d \mathbf{x}_t=\boldsymbol{f}\left(\mathbf{x}_t, t\right) d t+\boldsymbol{g}\left(\mathbf{x}_t, t\right) d \mathbf{w}_t, \quad t \in[0, T]
    \label{eq:forward_sde}
    \vspace{-0.5 em}
\end{equation}
where $\mathbf{w}$ is the standard Wiener process, $\boldsymbol{f}(\cdot)$ and $\boldsymbol{g}(\cdot)$ are scalar drift and diffusion coefficients, respectively. \citet{anderson1982reverse} states that the reverse process of \cref{eq:forward_sde} is also a diffusion process as:
\vspace{-0.5 em}
\begin{equation}
    d \mathbf{x}_s =\left[\boldsymbol{f}(\mathbf{x}_s, s)-\boldsymbol{g}(x_s, s)^2 \boldsymbol{s}\left(x_s, s\right)\right] \mathrm{d} s + \boldsymbol{g}(x_s, s) \mathrm{d} \mathbf{w}_s
    \label{eq:backward_sde}
\end{equation}
where $\mathbf{x}_s := x_{T-t}$ and $\boldsymbol{s}\left(x_s, s\right)$ := $\nabla_{\mathbf{x}_s} \log p_{s}\left(\mathbf{x}_s\right)$ is the score function of the marginal distribution over $x_s$.
\citet{song2020score} leverage this property to generate samples by first drawing $x_T \sim \mathcal{N}\left(\mathbf{0}, \mathbf{I}\right)$ and then solving the reverse SDE using a learned score network $\boldsymbol{s}_{\boldsymbol{\theta}}\left(\boldsymbol{x}_t, t\right)$.

By using Tweddie's formula \cite{stein1981estimation}, DDPM (Denoising Diffusion Probabilistic Model) \cite{ho2020denoising} can be shown as an equivalent interpretation of \cref{eq:backward_sde} \cite{luo2022understanding}:
\vspace{-0.5 em}
\begin{equation}
    \boldsymbol{s}_{\boldsymbol{\theta}}\left(\boldsymbol{x}_t, t\right) = - \frac{1}{\sqrt{1-\bar{\alpha}_t}} \boldsymbol{\epsilon}_{\boldsymbol{\theta}}\left(\boldsymbol{x}_t, t\right)
    \vspace{-1 em}
\end{equation}
where $\bar{\alpha}_t = \prod_{s=1}^t \alpha_s$, with mean coefficient $\alpha_t$, and $\boldsymbol{\epsilon}_{\boldsymbol{\theta}} \left(\boldsymbol{x}_t, t\right)$ is the noise network of DDPM.\\

\noindent\textbf{Definition of Visual Artifacts}
Generation flaws can be distinguished into two categories: Visual Artifacts and Hallucinations. Visual artifacts manifest as local irregularities or distortions in a generated image, such as blurred patches, unnatural textures, broken structures. In contrast, hallucinations refer to semantically generating incoherent content, such as extra limbs, misplaced objects or counterfactuals. In this paper, we focus specifically on detecting and correcting visual artifacts generated by diffusion models.

\begin{figure}[tb]
\centering
\includegraphics[width=0.95\linewidth]{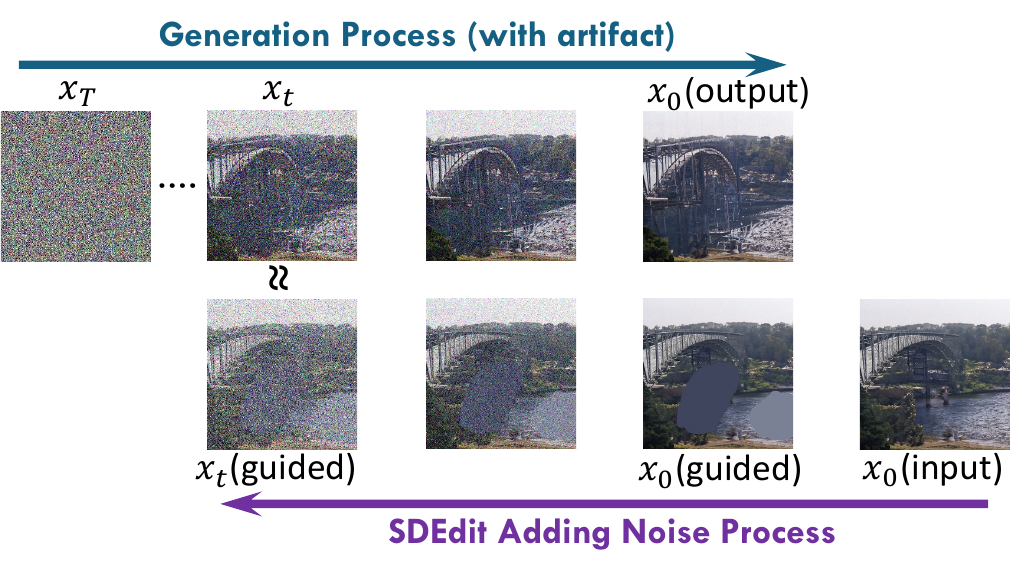}
\vspace{-1.5 em}
\caption{Artifact generation through denoising (top) and brush stroke noising via SDEdit \cite{meng2021sdedit} (bottom), demonstrating the model's inability to distinguish artifacts during generation.}
\vspace{-1.5 em}
\label{fig:sdedit_artifact}
\end{figure}

\begin{figure*}[htb]
\centering
\includegraphics[width=1\linewidth]{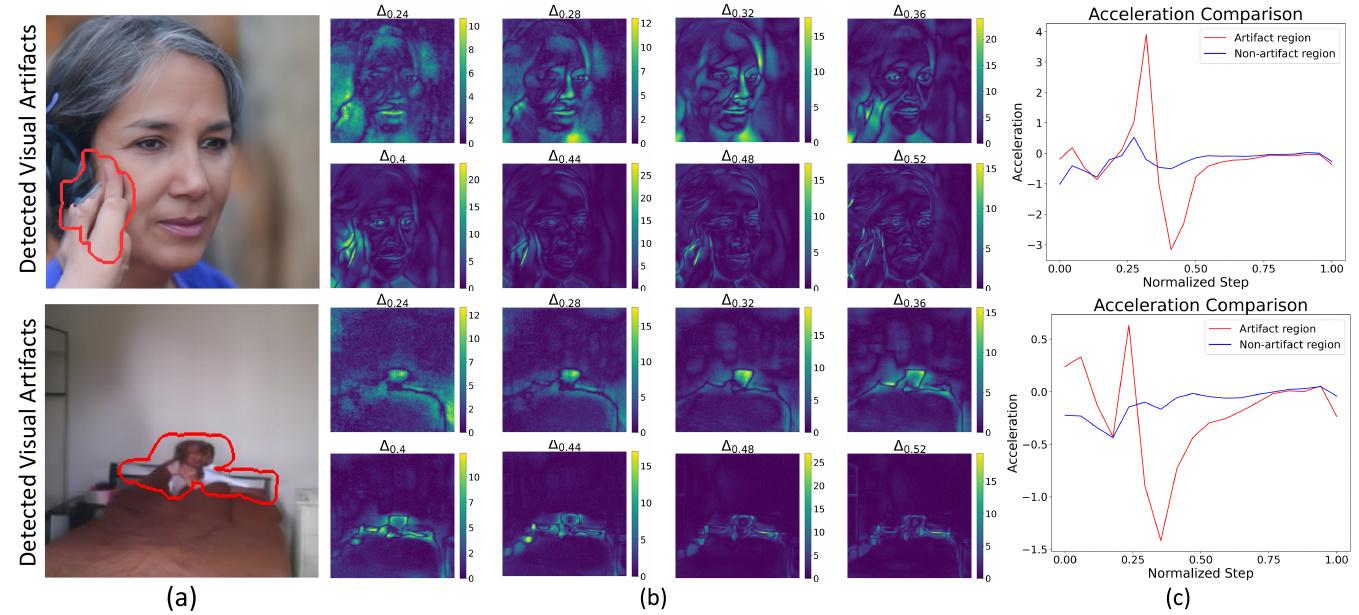}
\vspace{-2 em}
\caption{\textbf{Visualization of score dynamics and visual artifact detection.} \textbf{(a)} Generated images with detected visual artifact regions highlighted (red). \textbf{(b)} Visualization of score dynamics (normalized) between adjacent time steps as activation maps. Brighter regions (green to yellow) indicate areas of higher score variation, while darker regions (blue to black) show areas of lower score change. \textbf{(c)}. Score acceleration curves (representing the rate of change in score dynamics between consecutive timesteps) comparing artifact regions (red) with non-artifact regions (blue). The artifact regions exhibit characteristic rapid acceleration followed by deceleration, while non-artifact regions maintain stable score dynamics over time throughout a generative (inference) process.}
\label{fig:score_trap}
\vspace{-1 em}
\end{figure*}

\subsection{Detection by Anomalous Score Dynamic}
\label{sec:identification}

\noindent To understand how visual artifacts emerge during generations, we examine diffusion model behavior through the lens of image editing. SDEdit \cite{meng2021sdedit} demonstrates that diffusion models can transform irregularities, \eg, brush strokes, into semantically meaningful content through a noise-then-denoise process, revealing the inherent \textbf{Refinement} capability of diffusion models.
We observe that such irregularities after noising become structurally indistinguishable from states containing artifacts during generation, as shown in \cref{fig:sdedit_artifact}.
While diffusion models can successfully refine noised brush strokes, they fail to correct the corresponding artifacts during generation. This contrast reveals that diffusion models lack the ability to identify artifacts as patterns requiring refinement during the generation process.

To better understand this, we examine the generation process from a score perspective, as it directly represents the evolution of pixel values \cite{song2020score}. We define score dynamics as the difference between temporally adjacent score values: $\Delta s_\theta(x_t^{i, j}, t) = s_\theta(x_t^{i,j}, t) - s_\theta(x_{t-1}^{i,j}, t-1)$. 
Analysis reveals that image generation begins with establishing basic structures, followed by a phase of stochastic exploration where irregular patterns may emerge; we call these phases \textbf{Profiling} and \textbf{Mutation}, respectively. As shown in \cref{fig:score_trap}, regions containing visual artifacts exhibit characteristic patterns during mutation: They appear as localized regions of intense score variations (\cref{fig:score_trap}~(b)) and display dramatic acceleration followed by sudden deceleration in their score trajectories (\cref{fig:score_trap}~(c)). In contrast, normal regions maintain a stable evolution throughout generation.

Based on these observations, we propose a novel approach to detecting and localizing potential artifact regions over time in a diffusion inference process. Specifically, let $\Omega \subset \mathbb{R}^2$ denote the spatial domain of the image, and $\Omega_t^a \subset \Omega$ represent regions where abnormal evolution patterns emerge at timestep $t$. For each spatial location $(i,j) \in \Omega$, we track the score dynamics through consecutive timesteps. To account for the varying score magnitudes across different images and timesteps, we maintain a score bank $\mathcal{S} = \{s_\theta(x_k, k)\}_{k=t}^{T}$ and apply a temporal weighting function $w(t)$ that addresses the inherent decay of score magnitudes. Formally, we define artifact regions as:
\vspace{-0.5 em}
\begin{equation}
\Omega_t^a := \left\{(i,j) \in \Omega \mid \left| \Delta(w(t) \cdot s_\theta(x_t^{i,j}, t)) \right| > \tau \right\}
\label{eq:identification}
\vspace{-0.5 em}
\end{equation}
where $w(t) = \frac{1 - \bar{\alpha_t}}{\sqrt{\bar{\alpha_t}}}$ (see \cref{sec:theoretical} for theoretical analysis) and $\tau$ is adaptively determined as the maximum between the Median Absolute Deviation (MAD) of the weighted score dynamics and the mean of score bank $\mathcal{S}$. The final artifact regions $\Omega^a$ are accumulated across the score bank, with detailed procedures provided in \cref{algo:framework}.

\begin{algorithm*}[tbh]
\caption{Pseudo Code for Proposed \methodname~Method}
\begin{algorithmic}[1]
\State \textbf{Input:} Score network $\boldsymbol{s}_\theta(\cdot)$ which requires $T$ steps to generate, detection starting step $T_d$ and correction step $T_c$
\State \textbf{Initialize} $x_T \sim \mathcal{N}(0, \mathbf{I})$, Score Bank $\mathcal{S} \gets \{\}$, Visual Artifact Mask $\Omega^a \gets \{\}$ 
\For{$t$ = $T$, $t${-}{-}, while $t >= 0$}
    \State $\boldsymbol{x}_{t-1} = \sqrt{\alpha_{t-1}} \boldsymbol{x}_0 - \sqrt{1 - \alpha_{t-1}} \sqrt{1-\bar{\alpha}_t}\boldsymbol{s}_\theta(x_t, t)$, where $\boldsymbol{x}_0 = \frac{\boldsymbol{x}_t+\left(1-\bar{\alpha}_t\right) \boldsymbol{s}_\theta(x_t, t)}{\sqrt{\bar{\alpha}_t}}$ \hfill \(\triangleright\) Original Diffusion Process
    \If{$T_c< t <= T_d$} 
        \State $\mathcal{S}.\text{append}(\boldsymbol{s}_\theta(\boldsymbol{x}_t, t))$ \hfill \(\triangleright\) Store score value into Score Bank
    \ElsIf{$t == T_c$} \hfill \(\triangleright\) Anomalous Score Dynamics Detection Step
        \For{$k$ = 0, $k$++, while $k < (T_c - T_d)$} \hfill \(\triangleright\) Determine $\Omega^a$ by accumulation
        \State $\Omega^a = \Omega^a \cup \left\{(i,j) \in \Omega \mid \left| \Delta(w(k) \cdot s_\theta(x_k^{i,j}, k)) \right| > \tau \right\}$ \hfill \(\triangleright\) $\tau = \text{max} \{\text{MAD}(\Delta(w(k) \cdot s_\theta(x_k^{i,j}, k))), \text{mean}(\mathcal{S})\}$
        \EndFor
        \State $\boldsymbol{x}_t = \boldsymbol{x}_t \cdot \mathbbm{1}_{\overline{\Omega}^a}  + (\sqrt{\bar{\alpha}_t} \hat{\boldsymbol{x}}_0(t)+\sqrt{1-\bar{\alpha}_t} {\epsilon}) \cdot \gamma(t){\xi}\mathbbm{1}_{\Omega^a}$ \hfill \(\triangleright\) Trajectory-aware Targeted Correction Step
\EndIf
\EndFor
\State \textbf{return} $x_0$
\end{algorithmic}
\label{algo:framework}
\end{algorithm*}

\subsection{Real-Time Correction}
\label{sec:correction}
After detecting artifacts, we need a correction strategy to effectively refine the artifact region.
Two natural approaches emerge: either to revert to the states before the artifacts emerge through state replacement or to limit directly abnormal score changes through score clipping. For state replacement, we first predict the clean image from an earlier timestep $t$ using \cite{ho2020denoising}:
\vspace{-1 em}
\begin{equation}
\hat{\boldsymbol{x}}_0(\boldsymbol{x}_t, t) = \frac{1}{\sqrt{\bar{\alpha}_t}} \left(\boldsymbol{x}_t + (1-\bar{\alpha}_t) \nabla_\theta \log p(\boldsymbol{x}_t)\right),
\vspace{-0.5 em}
\label{eq:pred_xstart}
\end{equation}
Then replacing the artifact regions with corresponding states from this predicted clean image after re-noising to the current timestep.
Score clipping directly limits the magnitude of score changes during inference. 
However, both state replacement and score clipping fundamentally disrupt the mutation process, leading to reduced generation diversity. To address this problem, we propose Trajectory-aware Targeted Correction (TTC), which introduces controlled perturbations specifically in artifact regions at correction timestep $T_c$:
\begin{equation}
    \hat{\boldsymbol{x}}_{T_c} = \boldsymbol{x}_{T_c} \cdot \mathbbm{1}_{\overline{\Omega}^a}  + (\sqrt{\bar{\alpha}_{T_c}} \boldsymbol{x}_0^\prime+\sqrt{1-\bar{\alpha}_{T_c}} {\epsilon}) \cdot \gamma{\xi}\mathbbm{1}_{\Omega^a}
    \label{eq:ttc}
\end{equation}
where $\boldsymbol{x}_0^\prime = \hat{\boldsymbol{x}}_0(\boldsymbol{x}_{T_c}, T_c)$, ${\epsilon}, {\xi} \stackrel{\text{iid}}{\sim} \mathcal{N}(0, \mathbf{I})$, $\mathbbm{1}_{\Omega^a}$ is indicator function for artifact regions and perturbation intensity $\gamma$. 

TTC builds upon the understanding of score traps: Regions where pixels become locked in persistent score patterns after experiencing dramatic score changes. Through controlled perturbations, TTC disrupts these fixed patterns and allows pixels to resume normal evolution with surrounding regions. \cref{sec:theoretical} provides a detailed analysis of these score trap mechanisms and their relationship to visual artifacts. Generation quality and diversity comparison across correction strategies is shown in \cref{fig:diversity}.

\subsection{Theoretical Analysis}
\label{sec:theoretical}

We provide a theoretical analysis for the score trap and the choice of the temporal weighting function in \cref{eq:identification}.

\noindent\textbf{Theoretical Analysis of Score Traps}
For normal generation, the score evolution of each pixel is coupled with its neighborhood through the learned neural score function \cite{chen2018neural}:
\vspace{-0.5 em}
\begin{equation}
    s_\theta(x_t^{i,j}, t) = \nabla_{x_t^{i,j}} \log p_\theta(x_t^{i,j}|\mathcal{C}(i,j), t)
    \label{eq:score_traps}
\end{equation}
where $\mathcal{C}(i,j)$ represents the contextual information from neighboring pixels. This coupling ensures coordinated evolution toward the data manifold. When regions experience abnormal score dynamics, they can enter score traps where local patterns persist despite significant score values, disrupting the natural coupled evolution process. These trapped regions evolve based primarily on their local patterns, losing the contextual relationships necessary for coherent image generation.

This reveals how our perturbation-based correction re-establishes contextual relationships.  For a trapped pixel $(i,j)$, the perturbation $\gamma \cdot \xi$ introduces stochastic variations that disrupt the isolated score patterns, creating opportunities for these regions to re-couple with their surroundings through the natural score evolution process. Meanwhile, in areas without abnormal patterns, these modest perturbations preserve the original coupled evolution, ensuring the method remains harmless to non-artifact regions; 
see \cref{sec:harmless_nonartifact} for further mathematical derivations and proofs.\\

\begin{table*}[htb]
    \centering
    \vspace{-1 em}
    \caption{\textbf{Quantitative Comparisons} on five datasets. The methods compared include BayesDiff \cite{kou2023bayesdiff} and SARGD \cite{zheng2024self}, and three baseline methods: State Replacement Score Clipping and PAL \cite{zhang2023perceptual} + TTC. All methods use DDIM sampling with identical noise seeds to generate 10,000 images per dataset, ensuring each approach modifies the same deterministic trajectories for fair comparison. The best scores are in \textbf{bold} and second best in \underline{\textbf{underline with bold}}. Sup and UnS denote supervised and unsupervised methods, respectively.}
    \label{tab:comparison}
    \vspace{-1 em}
    \begin{adjustbox}{max width=\textwidth}
    \begin{tabular}{l|c|ccc|ccc|ccc|ccc|ccc}
    \Xhline{1.5 pt}
    \rowcolor{mygray}
     & & \multicolumn{3}{c|}{\textbf{FFHQ}\cite{karras2019style}} & \multicolumn{3}{c|}{\textbf{ImageNet}\cite{deng2009imagenet}} & \multicolumn{3}{c|}{\textbf{LSUN-Cat}\cite{yu2015lsun}} & \multicolumn{3}{c|}{\textbf{LSUN-Horse}\cite{yu2015lsun}} & \multicolumn{3}{c}{\textbf{LSUN-Bedroom}\cite{yu2015lsun}} \\
    \Xcline{3-5}{1.0 pt} \Xcline{6-8}{1.0 pt} \Xcline{9-11}{1.0 pt} \Xcline{12-14}{1.0 pt} \Xcline{15-17}{1.0 pt}
    \rowcolor{mygray}
    \multirow{-2}{*}{\textbf{Methods}} &
    \multirow{-2}{*}{\textbf{Type}} &
    \small{\textbf{FID} $\downarrow$} & \small{\textbf{Pre.} $\uparrow$} & \small{\textbf{Rec.} $\uparrow$} & \small{\textbf{FID} $\downarrow$} & \small{\textbf{Pre.} $\uparrow$} & \small{\textbf{Rec.} $\uparrow$} & \small{\textbf{FID} $\downarrow$} & \small{\textbf{Pre.} $\uparrow$} & \small{\textbf{Rec.} $\uparrow$} & \small{\textbf{FID} $\downarrow$} & \small{\textbf{Pre.} $\uparrow$} & \small{\textbf{Rec.} $\uparrow$} & \small{\textbf{FID} $\downarrow$} & \small{\textbf{Pre.} $\uparrow$} & \small{\textbf{Rec.} $\uparrow$} \\
    \hline \hline
    Original \cite{song2020denoising} & UnS & 36.69 & 0.629 & 0.493 & 14.68 & 0.739 & 0.734 & 22.17 & 0.513 & 0.586 & 29.36 & 0.510 & 0.642 & 12.96 & 0.627 & 0.583 \\
    State Replace & UnS & 37.09 & 0.635 & 0.495 & 14.61 & \textbf{\underline{0.743}} & 0.733 & 22.79 & 0.510 & 0.587 & 30.36 & 0.502 & 0.642 & 12.95 & 0.628 & 0.574 \\
    Score Clipping & UnS & 36.36 & 0.630 & 0.498 & 14.58 & 0.742 & \textbf{\underline{0.736}} & 22.12 & \textbf{\underline{0.515}} & 0.585 & 29.26 & 0.511 & 0.642 & 12.92 & 0.627 & \textbf{\underline{0.585}} \\
    \midrule
    BayesDiff \cite{kou2023bayesdiff} & UnS & 36.99 & 0.632 & 0.491 & 14.53 & \textbf{\underline{0.743}} & 0.730 & 22.50 & 0.513 & 0.585 & 28.70 & 0.518 & 0.634 & 12.88 & 0.625 & 0.569 \\
    SARGD \cite{zheng2024self} & Sup & 38.37 & \textbf{0.637} & 0.464 & 15.34 & 0.731 & 0.727 & 22.65 & \textbf{0.523} & 0.570 & 30.02 & 0.510 & 0.621 & 13.82 & \textbf{0.639} & 0.554  \\
    PAL \cite{zhang2023perceptual} + TTC & Sup & \textbf{\underline{36.35}} & 0.624 & \textbf{\underline{0.500}} & \textbf{14.01} & 0.731 & \textbf{0.747} & \textbf{21.83} & 0.514 & \textbf{\underline{0.588}} & \textbf{\underline{28.68}} & \textbf{\underline{0.519}} & \textbf{\underline{0.646}} & \textbf{\underline{12.71}} & \textbf{\underline{0.629}} & 0.579  \\ 
    \methodname~(Ours) & UnS & \textbf{36.28} & \textbf{0.637} & \textbf{0.503} & \textbf{\underline{14.41}} & \textbf{0.750} & 0.735 & \textbf{\underline{21.91}} & \textbf{\underline{0.515}} & \textbf{0.593} & \textbf{27.66} & \textbf{0.521} & \textbf{0.652} & \textbf{12.53} & 0.628 & \textbf{0.590} \\
    \bottomrule
    \end{tabular}
    \end{adjustbox}
    \vspace{-1.5 em}
\end{table*}

\noindent\textbf{Score Normalization}
The score function learned by diffusion models can be interpreted as a vector field guiding the denoising trajectory, inducing a probability flow \cite{song2020denoising}:
\vspace{-0.5 em}
\begin{equation}
    \frac{d\boldsymbol{x}}{dt} = -\frac{1}{2}\sigma_t^2 \nabla \log p_t(\boldsymbol{x}_t)
    \vspace{-0.5 em}
\end{equation}
The temporal evolution of this probability flow can be characterized by its divergence in the probability density field. Theoretically, we can model this through a flow operator $\mathcal{G}$:
\vspace{-0.5 em}
\begin{equation}
    \mathcal{G}(t) = \nabla \cdot \left(\frac{\partial}{\partial t} \int_{\tau=0}^t \mathcal{P}(\boldsymbol{x}_\tau, \tau) d\tau\right)
\end{equation}
where $\mathcal{P}(\boldsymbol{x}_\tau, \tau)$ denotes the local probability density at position $\boldsymbol{x}_\tau$ and time $\tau$. This formulation allows us to monitor the accumulation of probability density changes over time. We observe that the clean image prediction (\cref{eq:pred_xstart}) reflects these changes in the probability flow. Under the assumption of smooth probability density evolution between adjacent time steps \cite{sohl2015deep, kingma2021variational, song2020denoising}, score dynamics are captured through $\frac{\partial}{\partial t} \hat{\boldsymbol{x}}_0(t)$. Consequently, a normalization factor $w(t) = \frac{1 - \bar{\alpha_t}}{\sqrt{\bar{\alpha_t}}}$ is derived from the coefficient of the score term in \cref{eq:pred_xstart}, helping to equalize the scale of score variations throughout the denoising process.

\section{Experiments}

\noindent\textbf{Basic setups}
We conducted experiments on five datasets: FFHQ \cite{karras2019style}, ImageNet \cite{deng2009imagenet}, LSUN-Bedroom \cite{yu2015lsun}, LSUN-Cat \cite{yu2015lsun}, and LSUN-Horse \cite{yu2015lsun}. We employed the Guided Diffusion model framework and pre-trained weights from OpenAI \cite{dhariwal2021diffusion} and Segmentation-DDPM \cite{baranchuk2021label}. Quantitative evaluations were performed using FID \cite{heusel2017gans}, which measures the Fréchet distance between real and generated image distributions, along with Precision and Recall \cite{kynkaanniemi2019improved}, which evaluate sample fidelity and diversity, respectively.

\noindent\textbf{Implementation details}
For detecting diffusion artifacts, our approach was compared with LLaVA-v1.5-13B \cite{liu2024improved} and PAL \cite{zhang2023perceptual}, while artifact removal comparisons were made with BayesDiff \cite{kou2023bayesdiff} and the adapted SARGD \cite{zheng2024self}. 
All experiments were performed on NVIDIA A100 / H100 GPUs. We used DDIM \cite{song2020denoising} to improve inference efficiency with a Number of Function Evaluations (NFE) set to 25. \textit{\textbf{Remark}}: We demonstrate in \cref{sec:artifact_nfe} that there is no significant correlation between NFE and the generation of artifacts.
Full implementation details are provided in \cref{app:implementation}.

\subsection{Quantitative Comparisons to Existing Methods}
\label{sec:exp_correction}

We first evaluate the effectiveness of \methodname~in improving generative quality through comparisons with both unsupervised (BayesDiff \cite{kou2023bayesdiff}) and supervised (SARGD \cite{zheng2024self}) SOTA methods, along with the original diffusion model \cite{dhariwal2021diffusion} and two baselines from \cref{sec:correction} (state replacement and score clipping). To isolate the effectiveness of our correction method, we also evaluate a hybrid approach combining artifact detector PAL (used in SARGD) with our Trajectory-aware Targeted Correction (TTC). 
Quantitative results are shown in \cref{tab:comparison}. Among unsupervised methods, our \methodname~demonstrates superior performance across all datasets, consistently achieving better FID and Precision scores while maintaining higher Recall values than BayesDiff and baselines, indicating both improved generation quality and better preservation of diversity.

Compared to the supervised methods, our proposed \methodname~method shows leading performance across most experiments, achieving superior results on FFHQ, LSUN-Horse, and LSUN-Bedroom, while maintaining competitive performance on ImageNet and LSUN-Cat.
The better performances of PAL and SARGD on these two datasets are due to that they are supervised artifact detectors specifically trained on these datasets \cite{zhang2023perceptual}. Their advantages are not generalizable to other datasets FFHQ, LSUN-(Horse, Bedroom). In contrast, \methodname~as an unsupervised method has generalisable advantages in all domains without dataset specific training, making it more practical and scalable.
Additionally, our method demonstrates significant computational efficiency advantages. In our experiments, \methodname~detects and corrects artifacts in approximately 0.09\,s per image, which is 8.8$\times$ faster than PAL (0.79\,s).

The effectiveness of the correction mechanism (TTC) becomes particularly evident when comparing SARGD with PAL(used in SARGD) + TTC, where the trajectory-aware correction demonstrates significant advantages in preserving generation diversity across all datasets, reflected in consistently higher Recall scores. 

\begin{table}[htb]
\centering
\vspace{-0.5 em}
\caption{\textbf{Visual Artifact Detection Accuracy Comparison} between PAL (Supervised, Sup) \cite{zhang2023perceptual}, LLaVA (Zero-Shot, ZS) \cite{liu2024improved}, and our method (Unsupervised, UnS) on FFHQ \cite{karras2019style}, ImageNet \cite{deng2009imagenet}, and LSUN-(Bedroom, Cat, Horse) \cite{yu2015lsun}.}
\vspace{-1 em}
\label{tab:compare_llava}
\begin{adjustbox}{max width=\linewidth}
\begin{tabular}{l|c|ccccccc}
    \Xhline{1.5pt}
    \rowcolor{mygray}
    \textbf{Method} & \textbf{Type} & \textbf{FFHQ} & \textbf{ImageNet} & \textbf{Bedroom} & \textbf{Cat} & \textbf{Horse} \\
    \hline \hline
    PAL & Sup & 51.4\% & 69.2\% & 52.4\% & 69.8\% & 60.9\% \\
    LLaVA & ZS & 63.1\% & 91.1\% & 75.9\% & 59.5\% & 72.2\% \\
    Ours & UnS & 
    56.7\% \scriptsize{\textcolor{red}{(-6.4)}} & 
    67.7\% \scriptsize{\textcolor{red}{(-1.5)}} & 
    65.0\% \scriptsize{\textcolor{red}{(-10.9)}} & 
    68.3\% \scriptsize{\textcolor{red}{(-1.5)}} & 
    70.3\% \scriptsize{\textcolor{red}{(-1.9)}} \\
    \bottomrule\hline
\end{tabular}
\end{adjustbox}
\vspace{-1.5 em}
\end{table}

\begin{figure*}[htb]
\centering
\vspace{-1 em}
\includegraphics[width=1\linewidth]{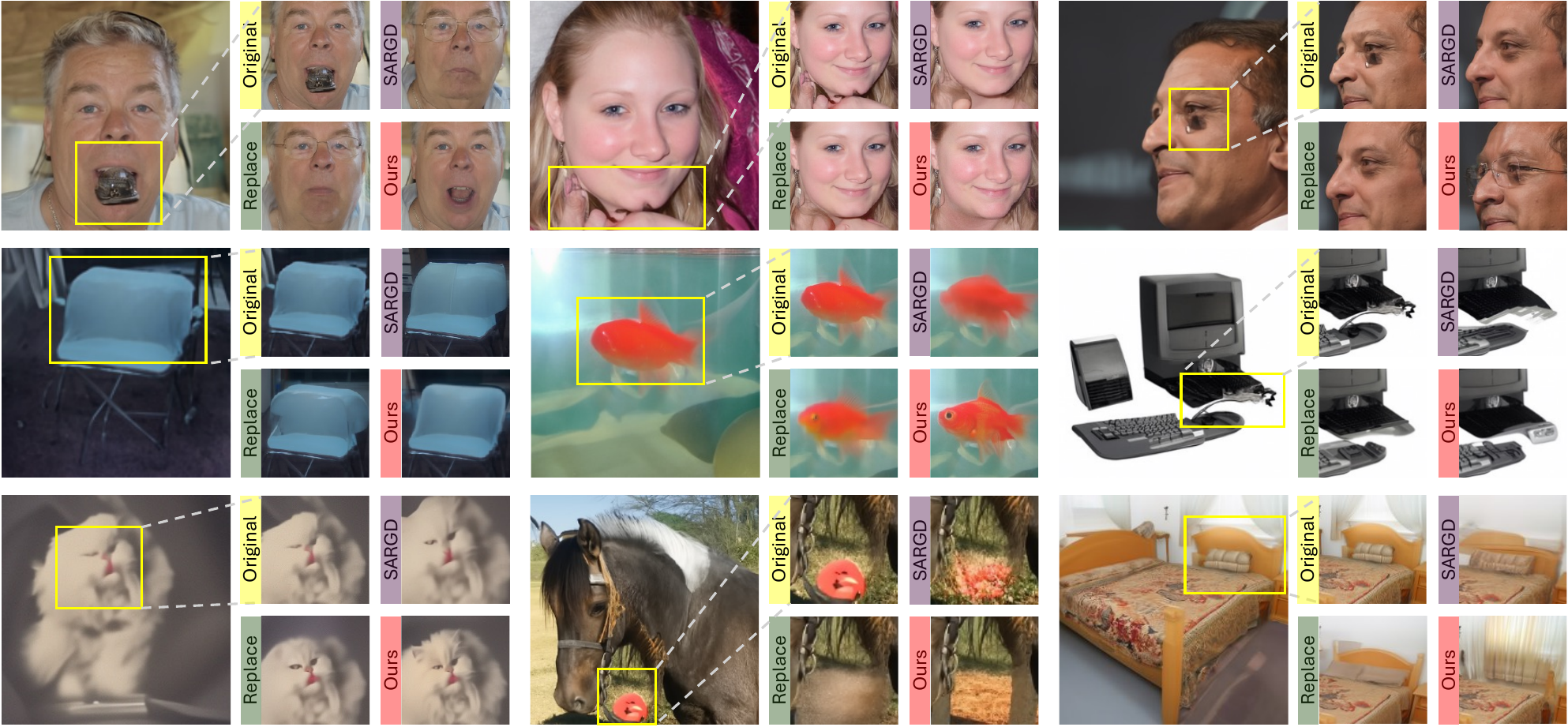}
\vspace{-2 em}
\caption{\textbf{Qualitative Comparison} of different correction methods. For each example, we show the original output with visual artifacts (left) and zoomed-in views of the artifact regions corrected by different methods (right): SARGD \cite{zheng2024self}, state replacement (Replace), and our trajectory-aware targeted correction (Ours). Rows from top to bottom: FFHQ\cite{karras2019style}, ImageNet\cite{deng2009imagenet}, and LSUN-(Cat, Horse, Bedroom)\cite{yu2015lsun}.}
\vspace{-1 em}
\label{fig:diversity}
\end{figure*}

\subsection{Artifact Detection Performance Analysis}
\label{sec:exp_identification}

To validate the accuracy of our method in identifying visual artifacts, we manually selected 200 images for each dataset from the diffusion model outputs, consisting of 100 images with visual artifacts and 100 without.
We then evaluated \methodname~against zero-shot large multi-modal model LLaVA-v1.5-13B \cite{liu2024improved} and supervised artifact detector PAL \cite{zhang2023perceptual}.
For LLaVA evaluation, we produced 50 different prompts and reported the results for the most effective prompt. Details on prompt generation are provided in \cref{sec:llava_prompts}. The accuracy for both methods is presented in \cref{tab:compare_llava}.

As an unsupervised method, \methodname~achieves promising detection performance, maintaining close accuracy to supervised approaches LLaVA and PAL across most datasets.
Notably, these methods analyze final generated images, whereas our approach detects artifacts during the generation process through score dynamics, enabling early intervention before artifacts fully manifest.
However, our method does show limitations in specific cases, such as low-contrast images where subtle abnormalities are difficult to distinguish from normal variations (leading to False Negatives), and instances where the diffusion model successfully rationalizes initially abnormal patterns during refinement (causing False Positives). Representative examples of these cases are illustrated in \cref{fig:failure}.

\subsection{Qualitative Analysis of Correction Methods}
To better illustrate the advantage of Trajectory-aware Targeted Correction (TTC) over baselines, \cref{fig:diversity} shows qualitative comparisons consisting of the original outputs with artifacts, results from state replacement (\cref{sec:correction}), SARGD \cite{zheng2024self}, and TTC (ours). 
While all methods can remove artifacts, TTC demonstrates superior detail preservation in corrected regions.
Specifically, both state replacement and SARGD tend to converge to similar expressions and local details, constraining natural variations, as they directly modify the generation states. SARGD faces further limitations from its artifact detector being trained on a specific domain \cite{zhang2023perceptual}, affecting its generalization ability.
More importantly, by preserving mutation phase operations, our trajectory-aware correction enables diverse yet coherent generations even when correcting the same region. Additional comparison results are provided in \cref{sec:more_corrected}.

\begin{figure}[htb]
\centering
\includegraphics[width=1\linewidth]{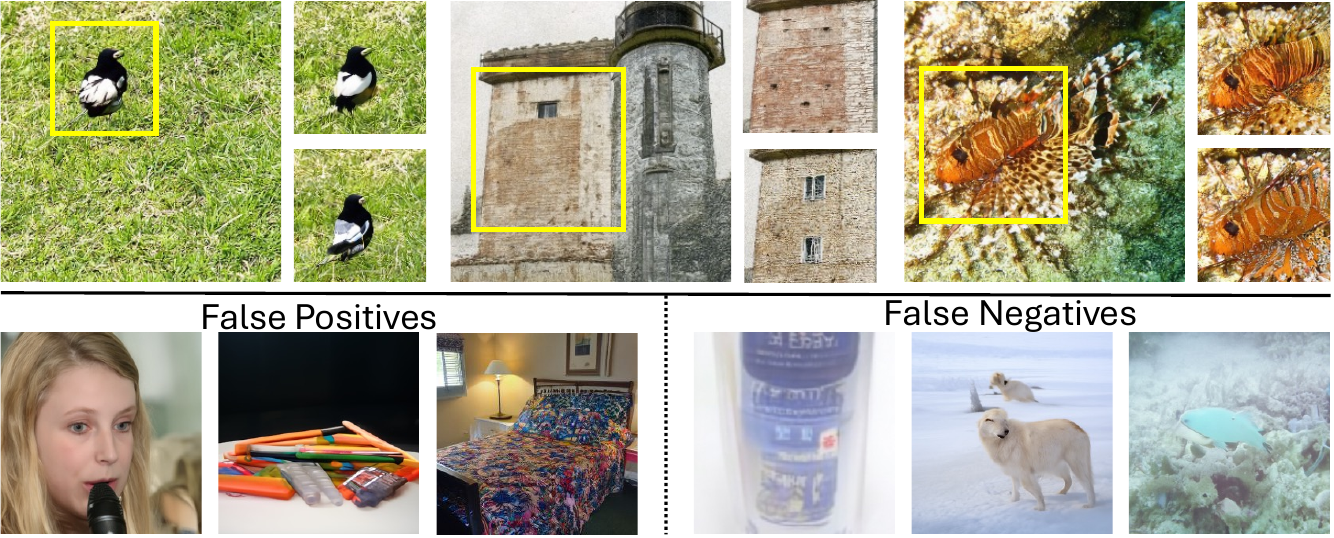}
\vspace{-2 em}
\caption{Top: Applied our correction method to clean regions (yellow box). Bottom: Typical failure cases.}
\vspace{-1.5 em}
\label{fig:failure}
\end{figure}

\noindent\textbf{Do corrections at Non-Artifact Regions harm?}
As with any detection method, our proposed scheme will inevitably encounter false positives, leading to add perturbations in non-artifact areas.
Our experiments demonstrate that applying the correction mechanism to regions without visual artifacts, as shown in \cref{fig:failure}, introduces modest variations while preserving semantic coherence with the surrounding context and not generating new artifacts.
Extended visual results and detection performance analysis can be found in \cref{sec:harmless_nonartifact} and \cref{sec:exp_identification}, respectively.

\subsection{Further Analysis}
\label{sec:further_analysis}

\begin{figure}[tb]
\centering
\includegraphics[width=1\linewidth]{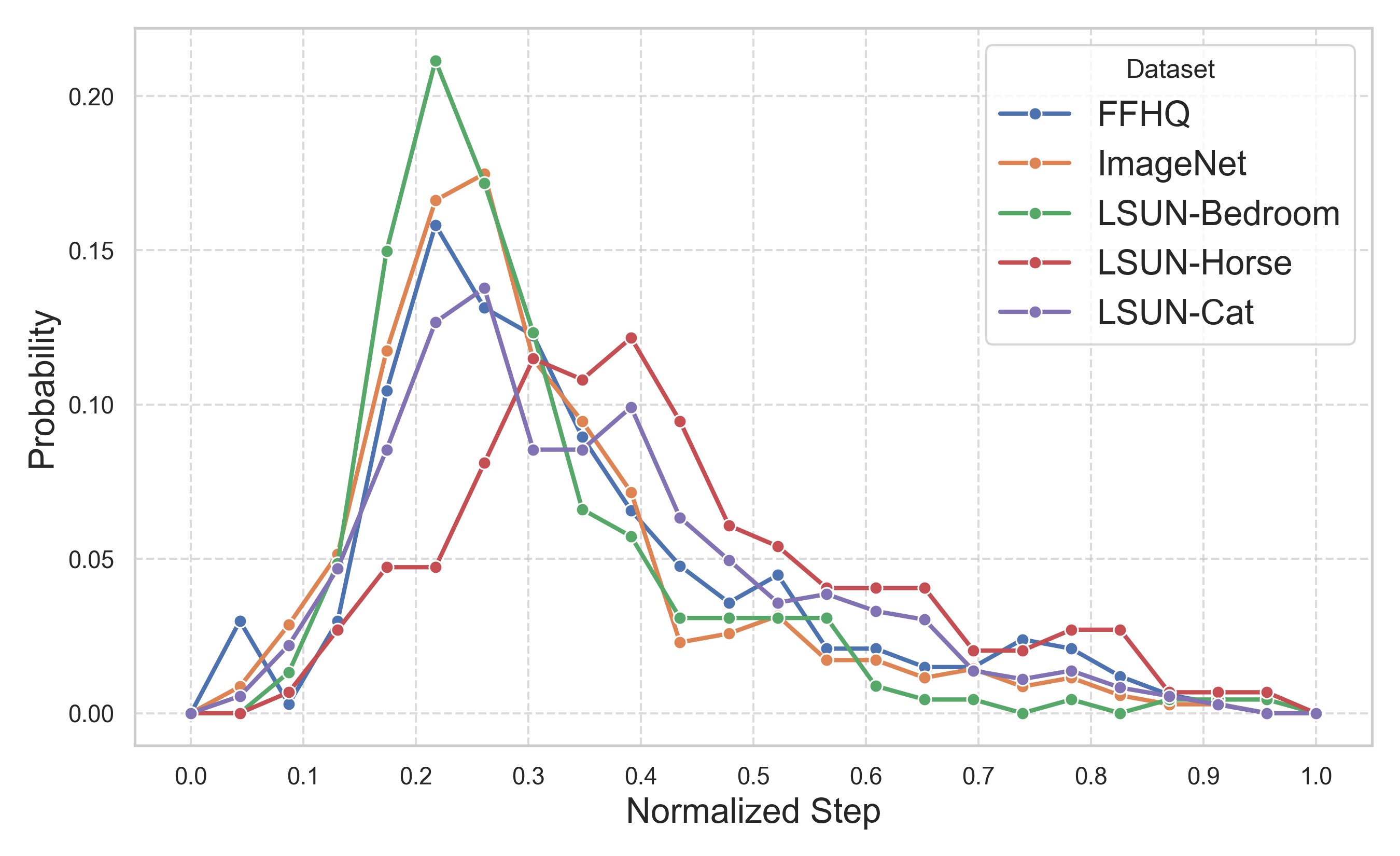}
\vspace{-8 mm}
\caption{Temporal Analysis of abnormal score dynamics across FFHQ \cite{karras2019style}, ImageNet \cite{deng2009imagenet}, LSUN-(Bedroom, Cat, Horse) \cite{yu2015lsun}.}
\vspace{-1.5 em}
\label{fig:artifact_distribution}
\end{figure}

\paragraph{Distribution of Abnormal Score Dynamics}
In \cref{fig:artifact_distribution}, we plot the frequency of abnormal scores at each time step (normalized by total diffusion time step $T$). The score dynamics demonstrate distinct patterns across different stages: remaining stable in early steps where basic structures emerge, experiencing significant variations in the middle stage, and gradually stabilizing again in the later steps during details' refinement. The presence of the long tail indicates that some variations persist into later steps, suggesting an extended period of details' adjustment. This behavioral pattern naturally aligns with our hypothesis that the generation undergoes three phases: profiling, mutation, and rationalization. This temporal pattern suggests that, while early intervention might be possible, determining the latest effective correction point is crucial for maximizing the detection of potential artifacts.\\

\begin{figure}[htb]
\centering
\vspace{-1 em}
\includegraphics[width=1\linewidth]{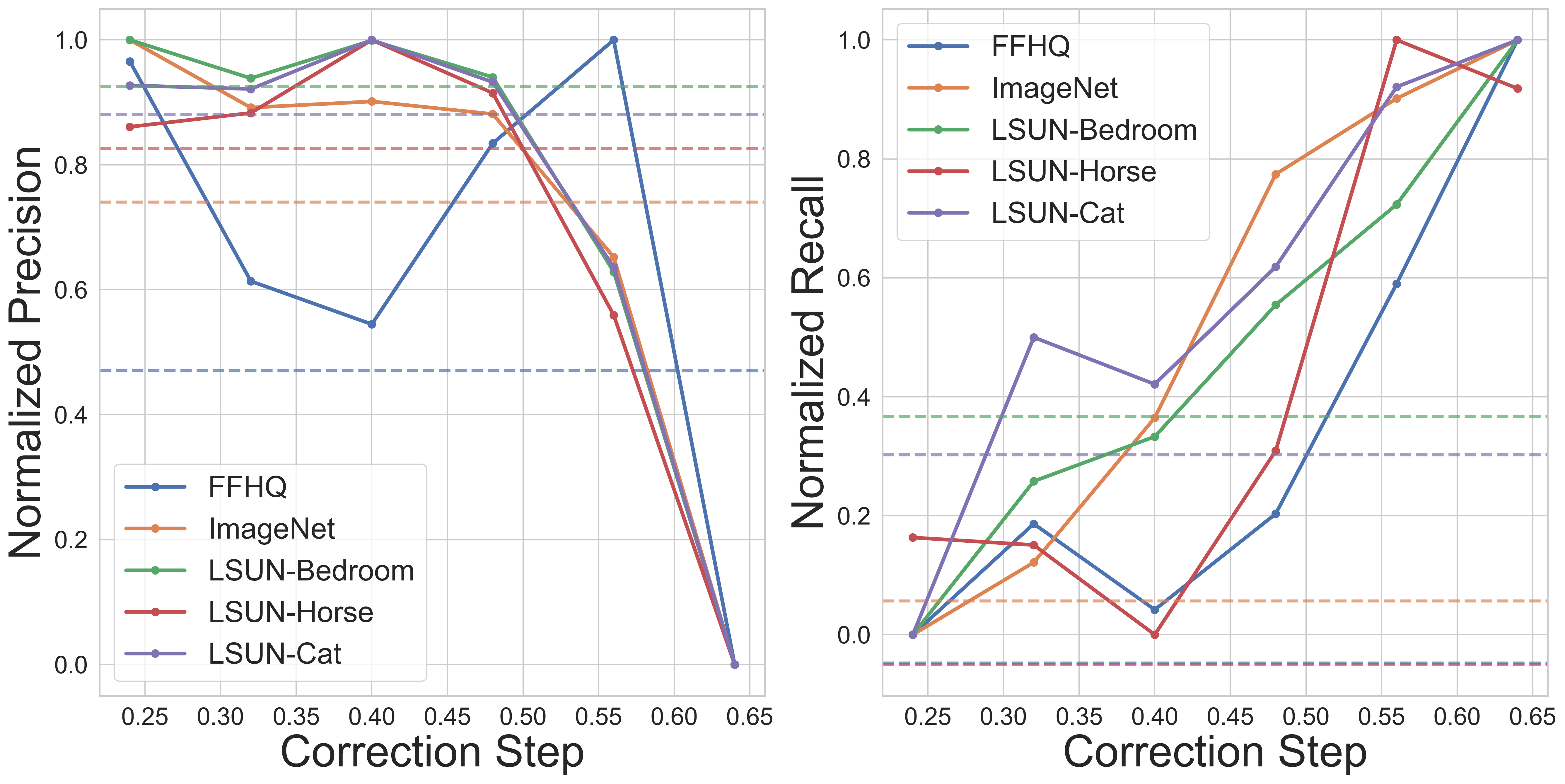}
\vspace{-2 em}
\caption{Impact of correction timestep $T_c$ on artifact removal performance evaluated by Precision (fidelity, $\uparrow$) and Recall (diversity, $\uparrow$). The dashed lines indicate the baseline precision / recall of the original diffusion model on each dataset.}
\vspace{-0.5 em}
\label{fig:correction_timing}
\end{figure}

\noindent\textbf{Impact of Correction Timing}
We investigate how the choice of correction timestep $T_c$ affects artifact removal effectiveness. Through extensive experiments, we identify a threshold at approximately $T_c^*/T \approx 0.48$ across different diffusion processes, representing the latest viable correction point before the model lacks sufficient steps for refinement. As shown in \cref{fig:correction_timing}, both Precision and Recall metrics achieve optimal performance as $T_c$ approaches $T_c^*$. This optimal timing allows for maximum artifact detection while ensuring adequate refinement steps.
Most datasets show stable performance before $T_c^*$ followed by a sharp decline, while FFHQ exhibits more fluctuation before $T_c^*$ and decreases gradually afterward. Individual dataset curves with analysis are provided in \cref{sec:individual_pr}.

\begin{figure}[htb]
\centering
\vspace{-1 em}
\includegraphics[width=1\linewidth]{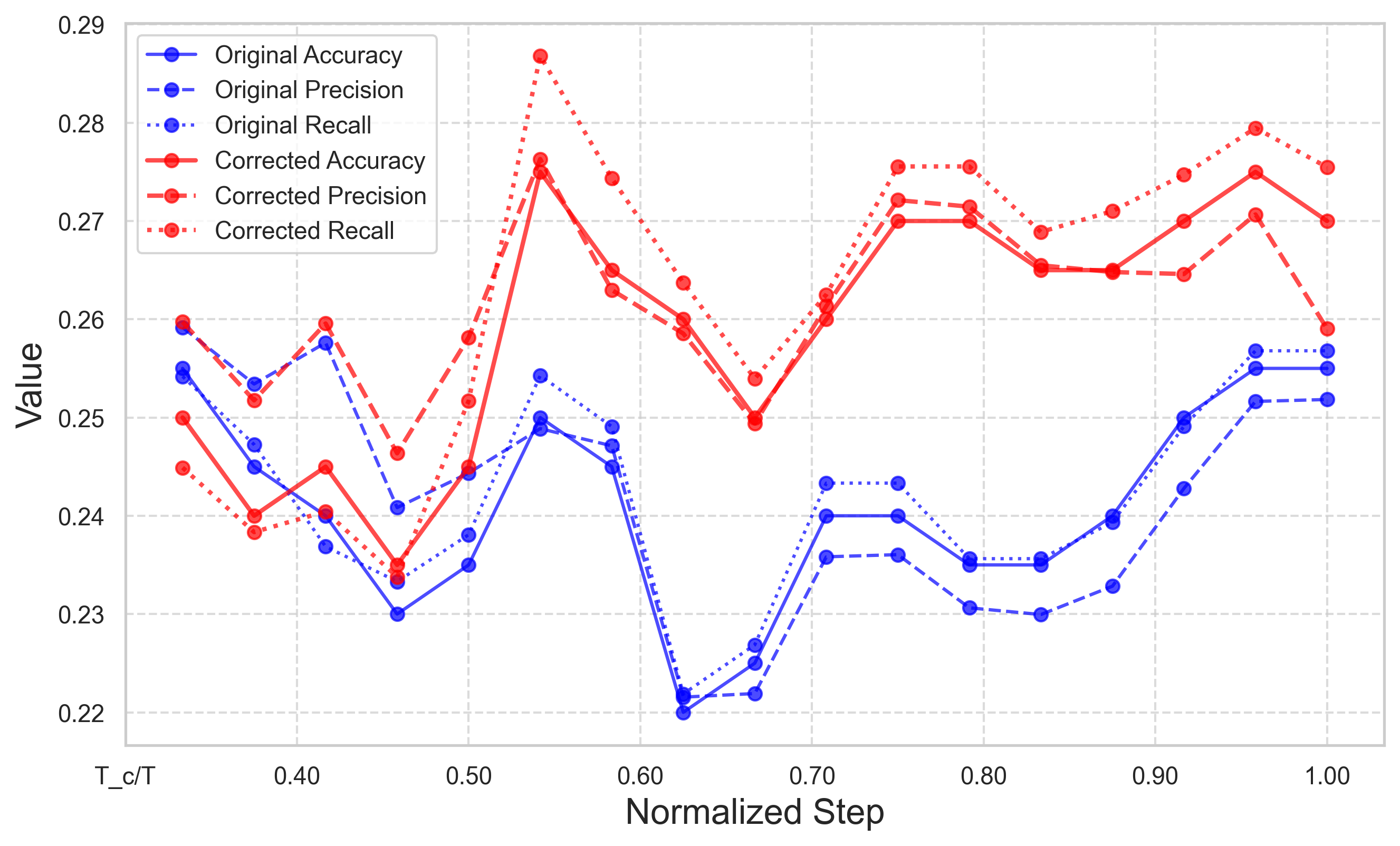}
\vspace{-2.5 em}
\caption{Linear probe classification results comparing latent representations from original and corrected diffusion trajectories across Accuracy, Precision and Recall metrics.}
\vspace{-1 em}
\label{fig:classification_imagenet}
\end{figure}

\noindent\textbf{Latent Code Improvement}
To evaluate how our correction method influences latent representations, we conduct a linear probe experiment \cite{xiang2023denoising} using a classifier-guided diffusion model \cite{dhariwal2021diffusion} on ImageNet \cite{deng2009imagenet}. 
Specifically, we generate samples following two paths: the original diffusion process and our corrected process. At an intermediate timestep $t$, we obtain the original state $x_t$ and apply our correction method to get the corrected state $\hat{x}_t$.
We then continue the diffusion process for $k$ steps to obtain $x_{t-k}$ and $\hat{x}_{t-k}$ from the original and corrected states, respectively. We generate $N$ labeled samples using both paths, with $y^i$ denoting the class label, resulting in two sets:
\begin{equation}
    \mathcal{D}_{\text{orig}} = \{(x_{t-k}^i, y^i)\}_{i=1}^{N}, \quad \mathcal{D}_{\text{corr}} = \{(\hat{x}_{t-k}^i, y^i)\}_{i=1}^{N}
\end{equation}
We train two separate classifiers on $\mathcal{D}_{\text{orig}}$ and $\mathcal{D}_{\text{corr}}$ respectively, with results shown in \cref{fig:classification_imagenet}. The higher accuracy achieved by improved latent codes throughout the remaining steps demonstrates that our method enhances the semantic quality of latent representations, and this improvement affects overall accuracy rather than individual precision or recall metrics. Notably, we observe that the classification accuracy reaches its peak earlier in the generation process with our method and remains stable through the refinement phase, while the original process exhibits a decrease-increase pattern during refinement.

\section{Conclusion}
We present a novel analysis of the diffusion generation process, decomposing it into profiling, mutation, and refinement phases, which provides fundamental insights into artifact formation mechanisms. Based on these insights, we develop \methodname, an unsupervised framework that successfully detects and corrects artifacts while preserving generation diversity. Extensive experiments demonstrate that \methodname~achieves competitive performance with state-of-the-art supervised methods across multiple datasets. The training-free nature of our approach enables immediate application to any diffusion model, making it a practical solution for improving generation quality.

\noindent\textbf{Future Work}
While our approach effectively detects artifacts through temporal pattern analysis, promising directions include improving detection in low-contrast regions, developing more robust discrimination between transient and persistent abnormalities, and extending these insights to other generative frameworks.

\section*{Acknowledgments} This work was partially supported by Veritone, Adobe, and has utilized Queen Mary’s Apocrita HPC facility from QMUL Research-IT.

{
    \small
    \bibliographystyle{ieeenat_fullname}
    \bibliography{main}
}

\clearpage
\maketitlesupplementary

\section*{Overview}
\label{app:overview}
This is the appendix for ``Temporal Score Analysis for Understanding and Correcting Diffusion Artifacts". \cref{tab:symbols} summarizes the abbreviations and symbols used in the paper.
This appendix is organized as follows:
\begin{itemize}
    \item Section \ref{app:implementation} presents additional implementation details, including the Pseudo-code of Adapted SARGD and the LLaVA prediction prompt generation method.
    \item Section \ref{app:analysis} presents an additional ablation study, including artifact persistence across NFE, harmless analysis, and more quantitative analysis.
    \item Section \ref{app:results} presents additional qualitative results, including more visualization of abnormal score dynamics (\cref{fig:more_abnormal_score_1}, \cref{fig:more_abnormal_score_2}), and corrected samples (\cref{fig:more_corrected_examples_1}, \cref{fig:more_corrected_examples_2}).
\end{itemize}

\begin{table}[htbp]
\centering
\caption{List of abbreviations and symbols used in the paper}
\vspace{-1 em}
\label{tab:symbols}
\begin{tabularx}{\linewidth}{lX}
\toprule
& \textbf{Meaning} \\
\midrule
\multicolumn{2}{l}{\textbf{Abbreviation}} \\
\methodname & Abnormal Score Correction for Enhancing Diffusion \\
TTC & Trajectory-aware Targeted Correction \\
DM & Diffusion Model \\
DDPM & Denoising Diffusion Probabilistic Model \\
NFE & Number of Function Evaluations \\
LMM & Large Multi-Modal Model \\
MAD & Median Absolute Deviation \\
\midrule
\multicolumn{2}{l}{\textbf{Symbol}} \\
$T_a$ & Artifact emerge step \\
$T_c$ & Artifact correction step \\
$T_c^*$ & Latest viable correction step \\
$T_d$ & Artifact detection starting step \\
$\Omega$ & Spatial location of an image \\
$\Omega_t^a$ & Artifact region at $t$ \\
$\Omega^a$ & Accumulated artifact region \\
$\mathcal{S}$ & Score Bank \\
$\tau$ & Adaptive abnormal score dynamic threshold \\
$\gamma$ & Perturbation intensity \\
$x_0$ & Final output Image from Diffusion Model \\
$x_0'$ & Predicted final output \\
$x_t$ & Intermediate state at $t$ \\
$w(t)$ & Temporal weighting function \\
$s_\theta(\cdot)$ & Score network \\
$\epsilon_\theta(\cdot)$ & Noise network \\
$T$ & Total time-steps \\
$\beta_1, \ldots, \beta_T$ & Variance schedule \\
$\alpha_t$ & $1 - \beta_t$ \\
$\bar{\alpha}_t$ & $\prod_{s=1}^t \alpha_s$ \\
\bottomrule
\end{tabularx}
\vspace{-2 em}
\end{table}

\section{Implementation Details}
\label{app:implementation}

\subsection{Adapted SARGD}
\label{sec:adapted_sargd}
Since SARGD \cite{zheng2024self} was originally designed for super-resolution where the final output (low resolution) is available, we adapt it to our scenario by using the predicted clean image ($x_0'$) at $T_d$ as guidance instead of the real Low-Resolution (LR) image. The rest of the correction process follows the original SARGD implementation, including artifact detection and refinement, but operates within our identified correction window ($T_c$ to $T_d$). The complete algorithm is provided in \cref{algo:sargd}, where the red mark-out text indicates removed steps from the original SARGD, and the green text shows our adaptations.

\begin{algorithm*}[htb]
\caption{Adapted Self-Adaptive Reality-Guided Diffusion (SARGD) Pseudo-code}
\label{algo:sargd}
\begin{algorithmic}[1]
\State \textbf{Input}: \textcolor{myred}{\sout{LR image $\boldsymbol{I}_{LR}$}}, and total diffusion steps $T$
\State \textbf{Load}: Encoder $\mathcal{E}$, artifact detector $\mathcal{A}$ and LR decoder $\mathcal{D}$
\State $\blacktriangleleft$ \textbf{Step 1: Initialization} \Comment{Removed as the final output $x_0$ is not accessible}
\State \textcolor{myred}{\sout{Upscale LR image as $up(\boldsymbol{I}_{LR})$}}
\State \textcolor{myred}{\sout{Encode the upsampled image as $\boldsymbol{x}=\mathcal{E}(up(\boldsymbol{I}_{LR}))$}}
\State \textcolor{myred}{\sout{Initialize the $\boldsymbol{x}$ as a realistic latent $\boldsymbol{x}_r$ and set it as guidance}}
\State \textcolor{myred}{\sout{Compute the realty score of the realistic latent $\boldsymbol{s}_r$}}
\State $\blacktriangleleft$ \textbf{Step 2: Sampling}
\For{t = T, ..., 1}
    \State Sample $\boldsymbol{\epsilon} \sim \mathcal{N}(\boldsymbol{0}, \boldsymbol{I})$ if $t > 1$, else $\boldsymbol{\epsilon} = 0$
    \State Computer the latent variable at the current step $\boldsymbol{x}_{t-1} = \frac{1}{\sqrt{\alpha_t}}\left(\boldsymbol{x}_t-\frac{1-\alpha_t}{\sqrt{1-\bar{\alpha}_t}} \epsilon_\theta\left(\boldsymbol{x}_t, \boldsymbol{x}, t\right)\right)+\sigma_\theta\left(\boldsymbol{x}_t, t\right) \boldsymbol{\epsilon}$
    \If{\textcolor{mygreen}{$t = T_d$}} \Comment{We use the predicted $x_0'$ to estimate the realty score}
        \State \textcolor{mygreen}{Set predicted $x_0'$ as guidance (using \cref{eq:pred_xstart}}
        \State \textcolor{mygreen}{Computer an estimated realty score of the realistic latent $\boldsymbol{s}_r$}
    \ElsIf{\textcolor{mygreen}{$T_c<= t < T_d$}} \Comment{Align SARGD correction timing with ours}
        \State \textbf{Detect artifacts} of the current latent $E_A=\mathcal{A}\left(\mathcal{D}\left(\boldsymbol{x}_{t-1}\right)\right)$ \Comment{Following steps remain the same}
        \State \textbf{Refine the latent} $\boldsymbol{x}_{t-1}=\boldsymbol{x}_{t-1} \times\left(1-E_A\right)+\boldsymbol{x}_r \times E_A$
        \State Decode the refined latent into an image $\mathbf{I}_r=\mathcal{D}\left(\boldsymbol{x}_{t-1}\right)$
        \State Generate the current binary reality map $M_R=\mathcal{R}\left(\mathbf{I}_r\right)$
        \State Calculate the current reality score $s_r^{t-1}=\mathcal{S}\left(M_R\right)$
        \State \textbf{Encode the current realistic latent} $\boldsymbol{x}_r^{t-1}=\mathcal{E}\left(\mathbf{I}_r\right)$
        \State \textbf{Update the guidance} $\boldsymbol{x}_r=\mathcal{G}\left(\boldsymbol{x}_r, \boldsymbol{x}_r^{t-1}\right)$ if $\boldsymbol{s}_r^{t-1}>\boldsymbol{s}_r$
        \State \textbf{Update the reality score} $\boldsymbol{s}_r=\boldsymbol{s}_r^{t-1}$ if $\boldsymbol{s}_r^{t-1}>\boldsymbol{s}_r$
    \EndIf
\EndFor
\State \Return the artifact-free \textbf{SR} $\boldsymbol{L}_{HR}=\mathcal{D}(\boldsymbol{x}_0)$
\end{algorithmic}
\end{algorithm*}

\subsection{LLaVA Prediction Prompts}
\label{sec:llava_prompts}
To generate effective prompts for LLaVA's \cite{liu2024improved} artifact detection, we first manually collected examples of images with artifacts. These examples were presented to GPT-4 \cite{achiam2023gpt} for prompt synthesis. We repeated this process 50 times, each time with different image combinations, generating 50 distinct prompts. The final evaluation used the best-performing prompt (No.~5) based on detection accuracy.
For reproducibility, we provide below the complete set of prompts used in \cref{tab:compare_llava}:

\begin{enumerate}
    \item ``Assess if there are any visible flaws in this image that a person could easily detect, like irregular shapes, unexpected color variations, blurred regions, or any other clear image disruptions. Answer with 'yes' or 'no'."
    
    \item ``Does this image contain any significant artifacts that distort the natural appearance, such as unexpected color patches, blurring, or pixelation? Please reply 'yes' or 'no'."
    
    \item ``Are there any obvious flaws in this image, such as large blurry areas, severe distortions, or color errors? Respond with 'yes' or 'no' only."
    
    \item ``Can you identify any glaring visual defects in this image that would be immediately noticeable to a human viewer? Reply with just 'yes' or 'no'."
    
    \item ``Determine if this image shows any noticeable defects or artifacts that would be easily seen by a human, including shape distortions, color issues, blurring, or pixelation in areas where it should be smooth. Please reply 'yes' or 'no'."
    
    \item ``Does this image have any obvious visual artifacts such as severe blurring, distortion, or unrealistic colors that would make it appear unnatural or of poor quality? Answer 'yes' or 'no'."
    
    \item ``Is the quality of this image significantly impaired by visual defects like large areas of pixelation, color mismatches, or misplaced objects? Please respond with 'yes' or 'no'."
    
    \item ``Can you identify any prominent visual issues in this image, such as incorrect color rendering, noticeable noise, severe blurring, or any elements that appear misplaced or distorted? Answer with 'yes' or 'no'."
    
    \item ``Does this image contain any obvious visual flaws that significantly degrade its quality, such as large blurry sections, strange artifacts, or clearly incorrect proportions of objects? Answer 'yes' or 'no'."
    
    \item ``Is there any obvious visual artifact in this image, like a hand growing out of a face, unrealistic color transitions, or large areas of texture inconsistency that make the image appear fake or unnatural? Please respond 'yes' or 'no'."
    
    \item ``Determine if this image has any clear visual artifacts that affect its appearance, such as distorted shapes, wrong color patches, excessive noise, or objects that are clearly in the wrong place. Reply with 'yes' or 'no'."
    
    \item ``Is the visual quality of this image compromised by obvious flaws, including but not limited to severe blurring, incorrect object placement, or large areas of unrealistic colors? Respond with 'yes' or 'no'."
    
    \item ``Examine the image for any significant visual issues, like pronounced noise, pixelation, unrealistic colors, or misaligned elements that affect the overall image quality. Please answer 'yes' or 'no'."
    
    \item ``Does this image exhibit any major visual artifacts that a human observer would immediately notice, such as large blurs, odd color patterns, or misplaced elements? Answer only with 'yes' or 'no'."
    
    \item ``Assess if there are any visible and distracting visual artifacts in this image, such as large unnatural blurs, obvious pixelation, incorrect object shapes, or areas of incorrect coloring. Reply with 'yes' or 'no'."
    
    \item ``Does this image contain any major visual artifacts that significantly degrade its quality? Answer only 'yes' or 'no'."
    
    \item ``Are there any obvious flaws in this image, such as large blurry areas, severe distortions, or color errors? Respond with 'yes' or 'no' only."
    
    \item ``Can you identify any glaring visual defects in this image that would be immediately noticeable to a human viewer? Reply with just 'yes' or 'no'."
    
    \item ``Does this image exhibit any significant visual anomalies like body parts in unnatural positions or severe pixelation? Answer 'yes' or 'no'."
    
    \item ``Is the overall quality of this image notably poor due to visible artifacts or distortions? Provide only a 'yes' or 'no' response."
    
    \item ``Are there any major visual imperfections in this image that make it look unrealistic or poorly generated? Reply with 'yes' or 'no'."
    
    \item ``Does this image contain any obvious flaws that would make you question its authenticity or quality? Answer only with 'yes' or 'no'."
    
    \item ``Can you spot any significant visual errors in this image, such as misplaced facial features or unnatural textures? Respond with just 'yes' or 'no'."
    
    \item ``Is there any clear evidence of poor image generation or editing in this picture, like inconsistent lighting or impossible anatomy? Reply 'yes' or 'no'."
    
    \item ``Does this image exhibit any significant visual anomalies like body parts in unnatural positions or severe pixelation? Answer 'yes' or 'no'." 
    
    \item ``Is the overall quality of this image notably poor due to visible artifacts or distortions? Provide only a 'yes' or 'no' response."
    
    \item ``Are there any major visual imperfections in this image that make it look unrealistic or poorly generated? Reply with 'yes' or 'no'."
    
    \item ``Does this image contain any obvious flaws that would make you question its authenticity or quality? Answer only with 'yes' or 'no'."
    
    \item ``Would you consider this image to be of low quality due to noticeable visual artifacts or errors? Answer with only 'yes' or 'no'."
    
    \item ``Does this image appear to be of normal quality, without any obvious visual artifacts such as blurring, distortion, or unnatural colors? Answer 'yes' if it appears normal, 'no' if there are visible issues."
    
    \item ``Is this image free of any significant visual defects like pixelation, color mismatches, or misplaced objects? Respond with 'yes' if there are no issues, or 'no' if such artifacts are present."
    
    \item ``Can you confirm that this image has no prominent visual issues, such as incorrect color rendering, noticeable noise, severe blurring, or misplaced elements? Answer 'yes' if there are no issues, and 'no' if there are."
    
    \item ``Can you spot any significant visual errors in this image, such as misplaced facial features or unnatural textures? Respond with just 'yes' or 'no'."
    
    \item ``Does this image lack any obvious visual flaws that would significantly degrade its quality, such as blurry sections, artifacts, or incorrect object proportions? Answer 'yes' for no flaws, 'no' if flaws are present."
    
    \item ``Is the image free from visual artifacts like hands growing out of faces, unrealistic color transitions, or texture inconsistencies that make the image look unnatural? Reply 'yes' if the image is clear, or 'no' if artifacts are present."
    
    \item ``Determine whether this image has any major visual artifacts affecting its appearance, such as distorted shapes, incorrect colors, excessive noise, or misaligned elements. Reply 'yes' if the image looks normal, or 'no' if such issues exist."
    
    \item ``Is the visual quality of this image high, with no obvious flaws like severe blurring, misplaced objects, or large patches of unrealistic colors? Respond 'yes' if the quality is good, 'no' if issues are found."
    
    \item ``Evaluate this image for any significant visual issues, such as noise, pixelation, unrealistic colors, or misaligned elements. Respond 'yes' if no issues are found, 'no' if any artifacts are present."
    
    \item ``Does this image have any noticeable visual artifacts that a human observer would immediately recognize, such as large blurs, odd color patterns, or misplaced elements? Answer 'yes' if there are no artifacts, or 'no' if artifacts are present."
    
    \item ``Is this image clear of any visible and distracting visual artifacts, such as large blurs, obvious pixelation, incorrect object shapes, or wrong coloring? Reply 'yes' if the image is free of artifacts, 'no' if artifacts are visible."
    
    \item ``Are there any jarring inconsistencies or unnatural elements in this image that detract from its realism? Answer 'yes' or 'no'."
    
    \item ``Does this image show any signs of poor rendering, such as incomplete objects or abrupt transitions? Respond with only 'yes' or 'no'."
    
    \item ``Can you detect any major issues with perspective or proportions in this image that make it look artificial? Reply with 'yes' or 'no'."
    
    \item ``Are there any noticeable problems with the lighting or shadows in this image that seem unrealistic? Answer only 'yes' or 'no'."
    
    \item ``Does this image contain any elements that appear to be unnaturally distorted or warped? Provide a 'yes' or 'no' response."
    
    \item ``Can you identify any significant issues with the texture or surface details in this image that look artificial? Reply with just 'yes' or 'no'."
    
    \item ``Are there any obvious problems with the edges or outlines of objects in this image, such as jagged lines or haloing? Answer 'yes' or 'no'."
    
    \item ``Does this image exhibit any clear signs of over-processing or artificial enhancement that degrade its quality? Respond with 'yes' or 'no' only."
    
    \item ``Can you spot any major inconsistencies in the style or appearance of different parts of this image? Reply with 'yes' or 'no'."
    
    \item ``Are there any glaring issues with the color balance or saturation in this image that make it look unnatural? Answer only with 'yes' or 'no'."
\end{enumerate}

\begin{figure*}[htb]
\centering
\vspace{-1 em}
\includegraphics[width=1\linewidth]{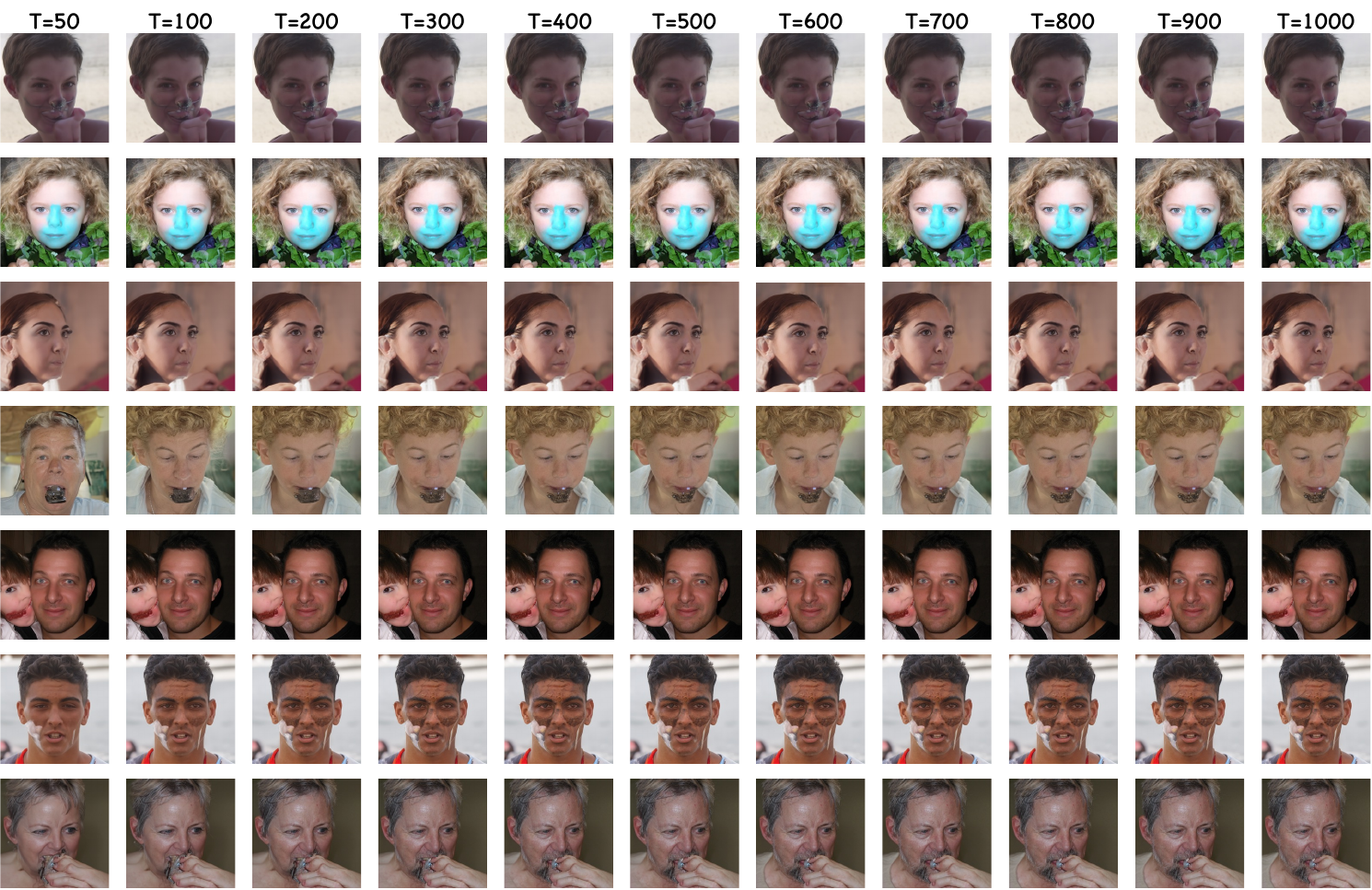}
\vspace{-2 em}
\caption{Visual artifacts persist across different numbers of sampling steps (NFE (T) from 50 to 1000 (original)) using the same random seed. While non-artifact regions show minor evolution in details, artifact regions remain virtually unchanged, demonstrating that artifacts stem from disrupted score dynamics rather than insufficient sampling granularity.}
\vspace{-1 em}
\label{fig:larger_nfe}
\end{figure*}

\section{Additional Analysis}
\label{app:analysis}

\subsection{Artifact Persistence Across NFE}
\label{sec:artifact_nfe}
Our main experiments use DDIM with NFE$=25$ for efficiency. To evaluate the potential effects of sampling granularity, we tested increasing NFE by up to 1000 steps (original).
As shown in \cref{fig:larger_nfe}, while a higher NFE allows more iterations for pixel evolution, leading to changes in overall image composition, the artifact regions remain visually unchanged. This observation supports our score trap analysis (\cref{sec:theoretical}): surrounding pixels continue to evolve with more sampling steps, but the trapped regions maintain their patterns, demonstrating that these areas indeed stop updating due to disrupted score dynamics rather than insufficient sampling steps.

\begin{figure*}[htb]
\centering
\includegraphics[width=1\linewidth]{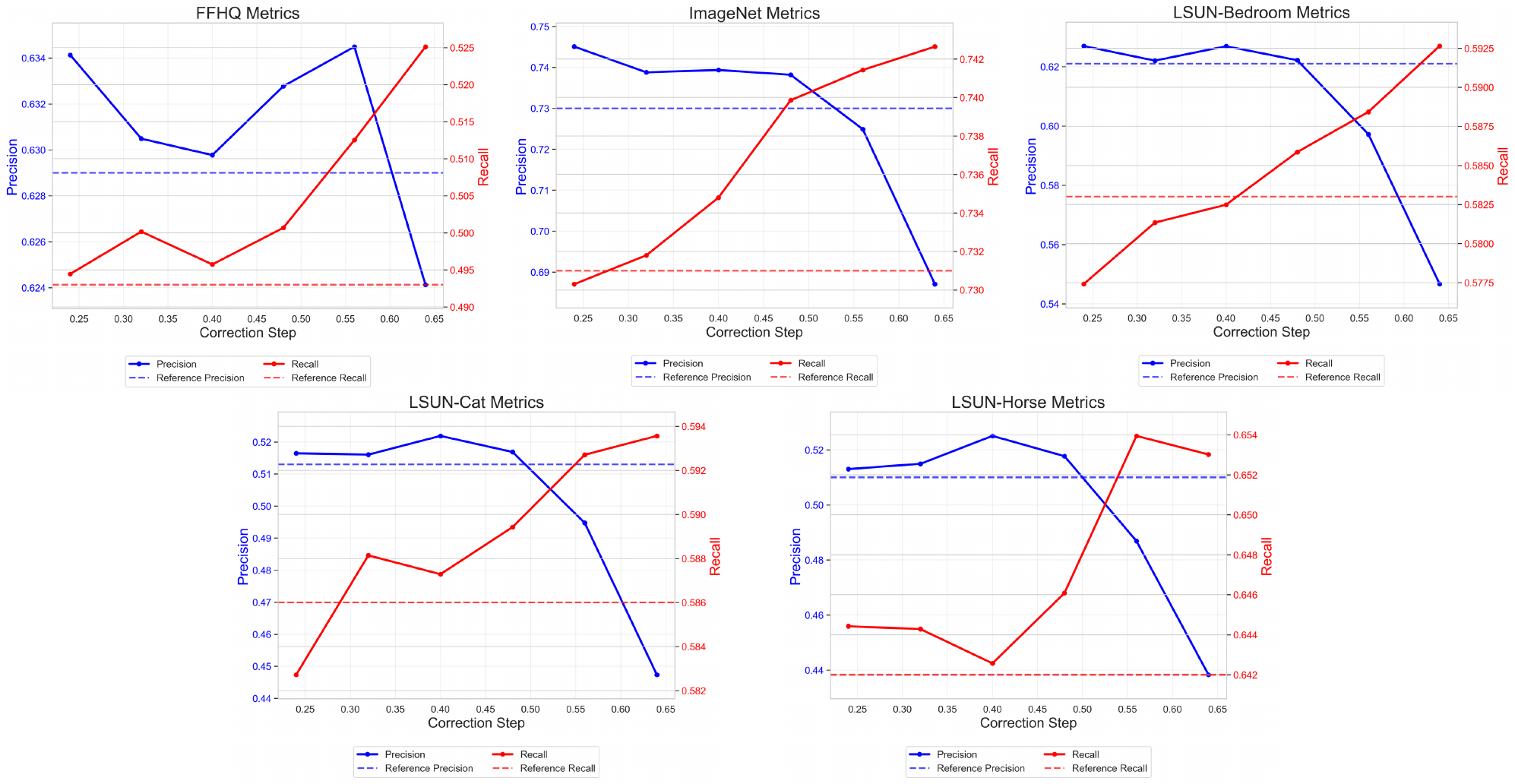}
\vspace{-2 em}
\caption{Impact of correction timestep $T_c$ on artifact removal performance for FFHQ, ImageNet, LSUN bedrooms, horses, and cats. For each dataset, the blue and red solid lines show the Precision (fidelity) and Recall (diversity), respectively, while the corresponding dashed lines indicate the baseline performance of the original diffusion model.}
\vspace{-1 em}
\label{fig:individual_metric}
\end{figure*}

\subsection{Harmlessness in Non-Artifact Regions}
\label{sec:harmless_nonartifact}
To understand why our correction method maintains semantic coherence while enabling controlled diversity in non-artifact regions, we need to examine both the local score dynamics and the fundamental properties of diffusion models. In normal regions, pixels maintain coupled evolution through the score function as described in \cref{eq:score_traps}, where each pixel evolves in coordination with its neighborhood context. When our method introduces controlled perturbations in these regions, two mechanisms work in concert to preserve image integrity.

\begin{figure*}[htb]
\centering
\includegraphics[width=0.90\linewidth]{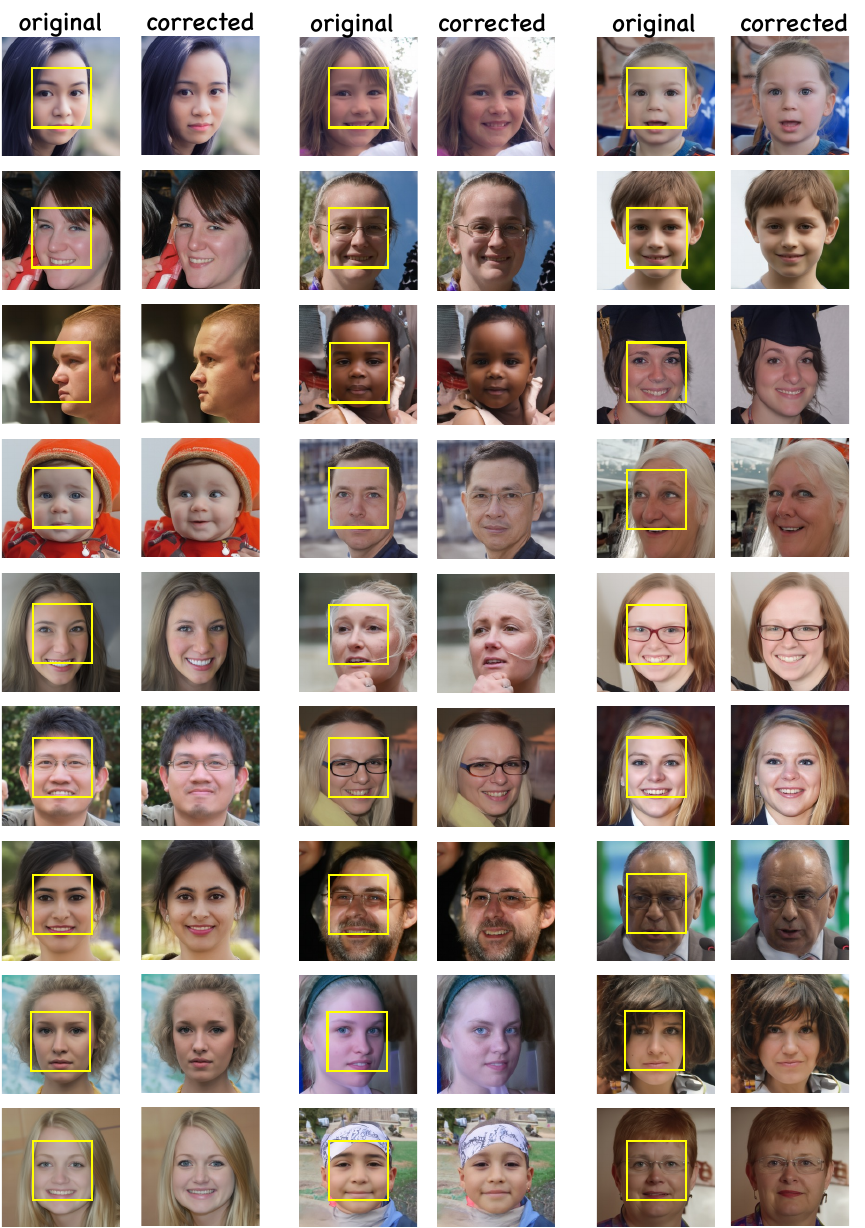}
\vspace{-1 em}
\caption{Applied our correction method (Trajectory-aware Targeted Correction) to clean region (yellow box).}
\vspace{-1 em}
\label{fig:more_safety}
\end{figure*}

First, the coupled score evolution pattern remains intact, as these regions maintain normal dynamics without entering score traps. This coupling naturally guides the perturbed pixels to evolve in harmony with their surroundings. Second, and more fundamentally, diffusion models are inherently equipped to handle noise through their denoising objective:
\vspace{-1 em}
\begin{align}
&\underset{\boldsymbol{\theta}}{\arg \min } 
D_{\mathrm{KL}}\left(q\left(\boldsymbol{x}_{t-1} \mid \boldsymbol{x}_t, \boldsymbol{x}_0\right) \, \middle| \, p_{\boldsymbol{\theta}}\left(\boldsymbol{x}_{t-1} \mid \boldsymbol{x}_t\right)\right)\\
= & \underset{\boldsymbol{\theta}}{\arg \min }\frac{1}{2 \sigma_q^2(t)} \frac{\bar{\alpha}_{t-1}\left(1-\alpha_t\right)^2}{\left(1-\bar{\alpha}_t\right)^2}\left[\left\|\hat{x}_{\boldsymbol{\theta}}\left(\boldsymbol{x}_t, t\right)-\boldsymbol{x}_0\right\|_2^2\right]
\label{eq:objective}
\end{align}
where $\hat{x}_\theta(\cdot)$ predicts $x_0$ directly \cite{kingma2021variational}. Following \cite{luo2022understanding}, this objective can be rewritten in terms of Signal-to-Noise Ratio (SNR):
\vspace{-1 em}
\begin{equation}
    \underset{\boldsymbol{\theta}}{\arg \min } \frac{1}{2}(\operatorname{SNR}(t-1)-\operatorname{SNR}(t))\left[\left\|\hat{\boldsymbol{x}}_{\boldsymbol{\theta}}\left(\boldsymbol{x}_t, t\right)-\boldsymbol{x}_0\right\|_2^2\right]
\label{eq:snr}
\end{equation}
where $\operatorname{SNR}(t)=\frac{\bar{\alpha}_t}{1-\bar{\alpha}_t}$. This formulation reveals that the diffusion process naturally increases SNR during denoising, ensuring controlled perturbations are effectively processed while maintaining semantic structure through coupled score evolution. Additional visual examples of perturbation effects in non-artifact regions are provided in \cref{fig:more_safety}.

\subsection{Analysis of Global Correction Application}
Given our perturbations maintain semantic coherence and introduce controlled diversity in non-artifact regions, a natural question arises: Why not extend these perturbations to the entire image regardless of artifact detection? When applying correction globally, each image would essentially undergo a ``second" generation process. Since the underlying diffusion model has an inherent probability of generating artifacts, this universal application would maintain the same artifact rate rather than reduce it. Therefore, introducing perturbations selectively based on artifact detection is necessary, avoiding unnecessary variations in well-formed regions while preserving the diversity benefits where needed.

\subsection{Individual Precision and Recall}
\label{sec:individual_pr}
Following the timing analysis in the main text (\cref{sec:further_analysis}), we present detailed performance curves for each dataset in \cref{fig:individual_metric}. The results reveal different patterns across datasets: while most datasets exhibit a clear performance drop after $T_c^*$, FFHQ shows a more gradual degradation. This difference can be attributed to the complexity of facial features, which allows for more flexible refinement compared to other domains. Notably, all datasets maintain performance above their respective baselines (shown in dashed lines) when corrections are applied before $T_c^*$, demonstrating the robustness of the identified threshold. The consistent pattern of optimal performance near $T_c^* \approx 0.48$ in various datasets validates the generality of this timing criterion for diffusion artifact correction.

\section{Additional Experiment Results}
\label{app:results}

\subsection{More Abnormal Score Dynamics Visualization}
\begin{figure*}[htb]
\centering
\includegraphics[width=0.8\linewidth]{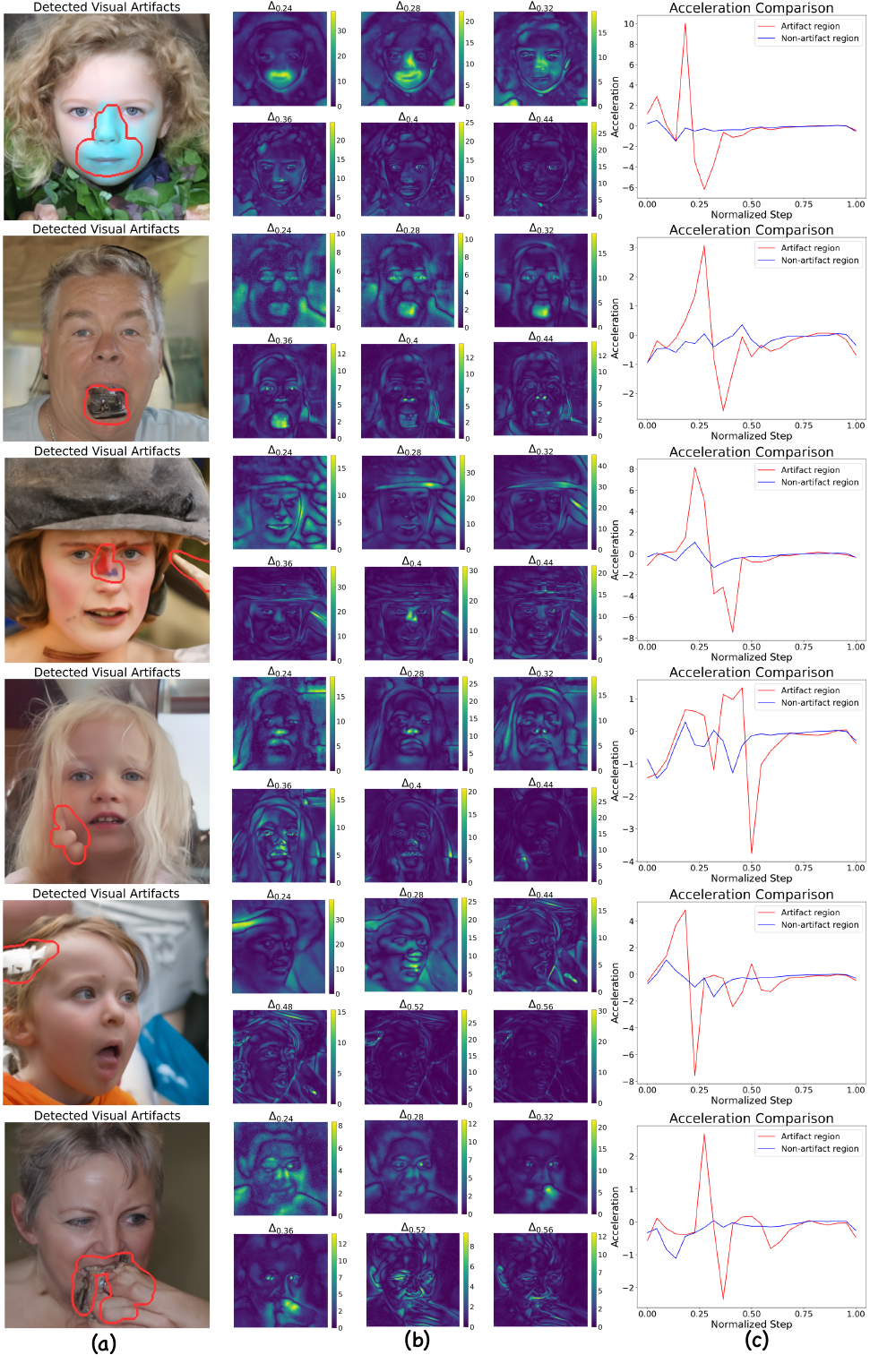}
\vspace{-1 em}
\caption{Extended visualization of abnormal score dynamics and visual artifact detection with more examples, following the analysis shown in \cref{fig:score_trap} of the main text. The same patterns of score acceleration and deceleration in artifact regions are consistently observed across different cases.}
\vspace{-2 em}
\label{fig:more_abnormal_score_1}
\end{figure*}

\begin{figure*}[htb]
\centering
\includegraphics[width=0.8\linewidth]{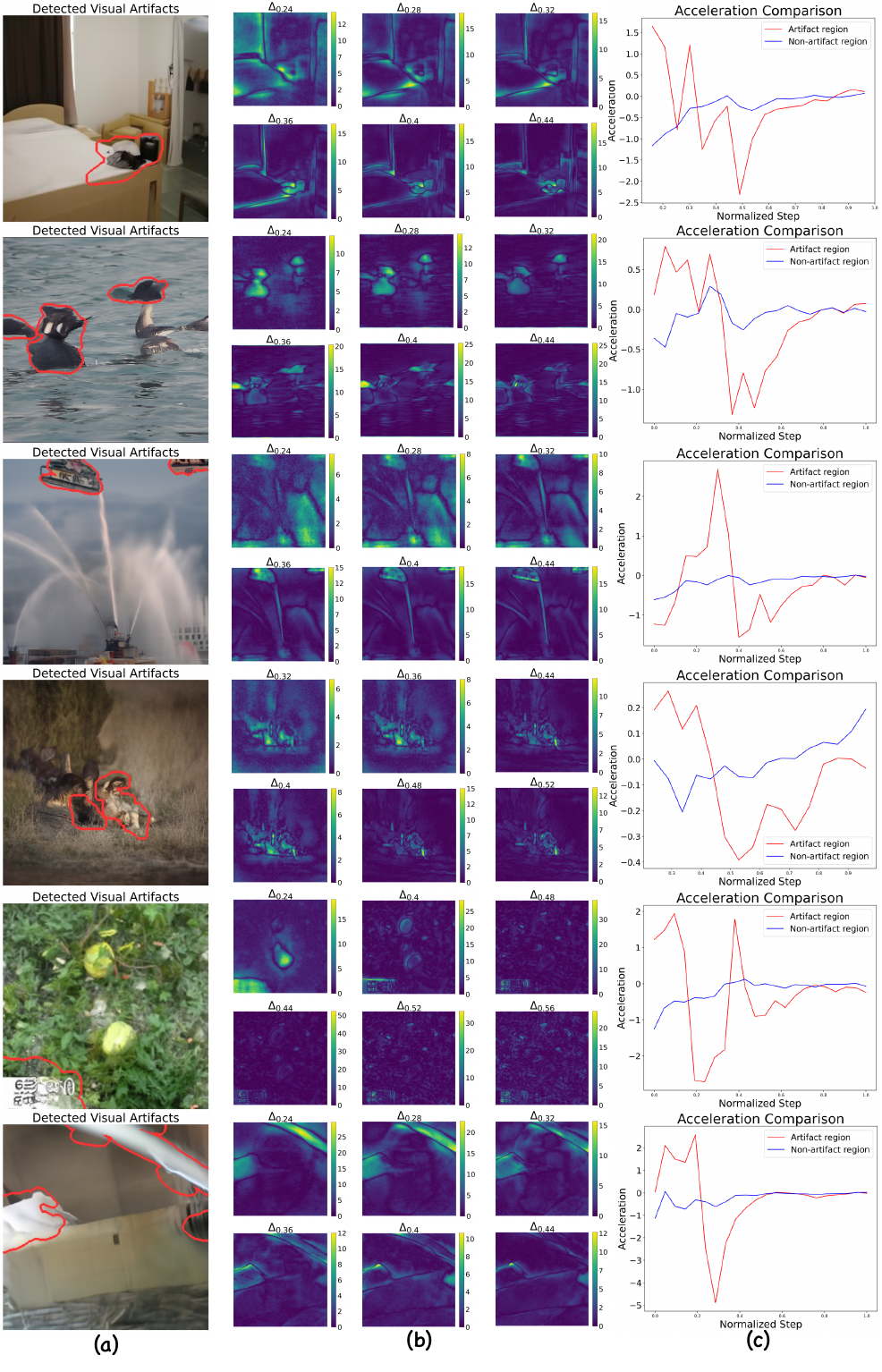}
\vspace{-1 em}
\caption{Extended visualization of abnormal score dynamics and visual artifact detection with more examples, following the analysis shown in \cref{fig:score_trap} of the main text. The same patterns of score acceleration and deceleration in artifact regions are consistently observed across different cases.}
\vspace{-2 em}
\label{fig:more_abnormal_score_2}
\end{figure*}

Extended from the representative cases in the main paper (\cref{fig:score_trap}), \cref{fig:more_abnormal_score_1} and \cref{fig:more_abnormal_score_2} show additional qualitative analysis of score dynamics in artifact regions. These examples consistently demonstrate the characteristic abnormal score patterns: sharp variations in score changes displayed in activation maps and the distinct acceleration-deceleration curves in artifact regions compared to normal areas.

\subsection{More Corrected Samples}
\label{sec:more_corrected}
\begin{figure*}[htb]
\centering
\includegraphics[width=0.85\linewidth]{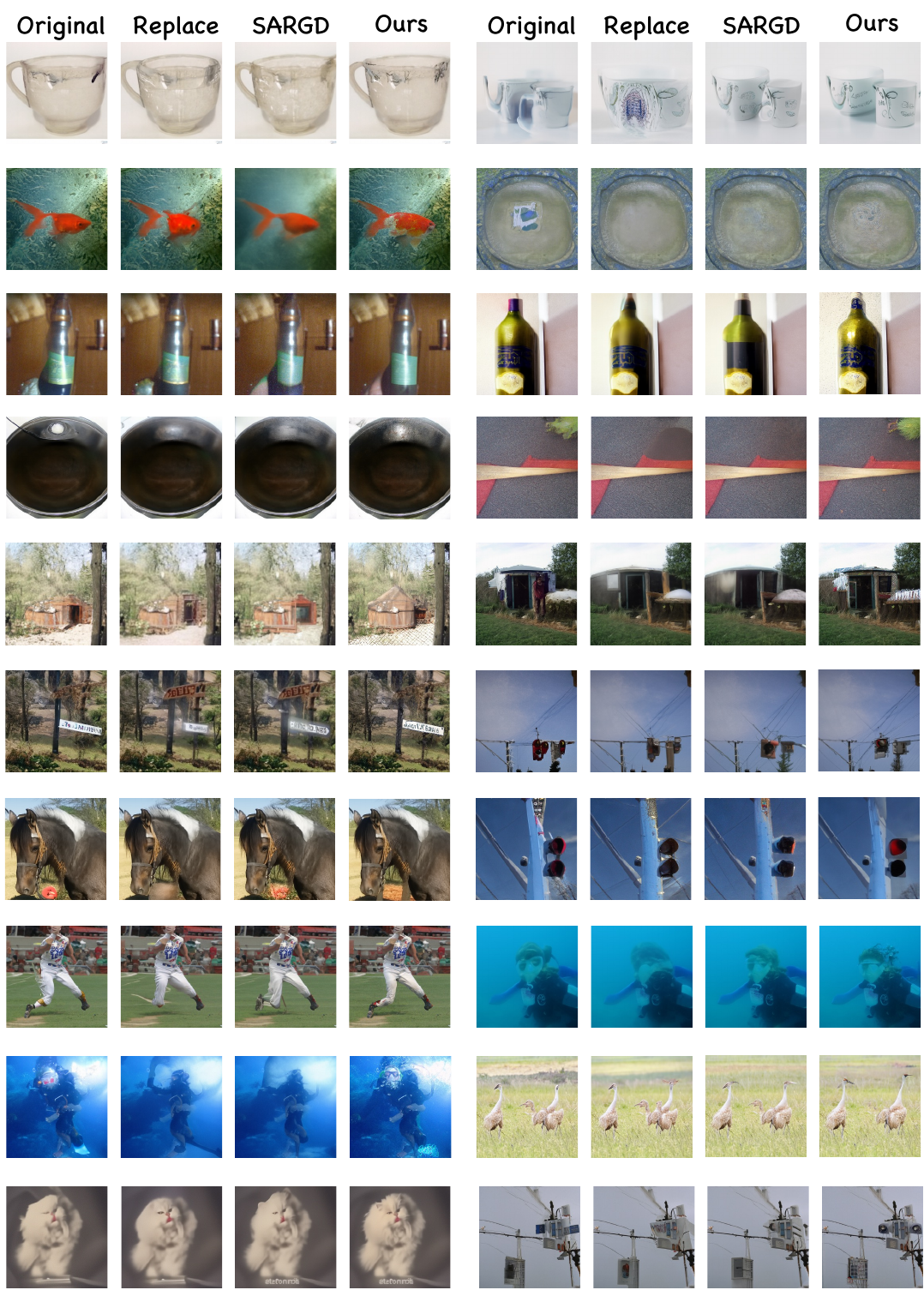}
\vspace{-1 em}
\caption{Additional qualitative comparison of artifact correction methods, following the similar format as \cref{fig:diversity} in the main text.}
\vspace{-1 em}
\label{fig:more_corrected_examples_1}
\end{figure*}

\begin{figure*}[htb]
\centering
\includegraphics[width=0.85\linewidth]{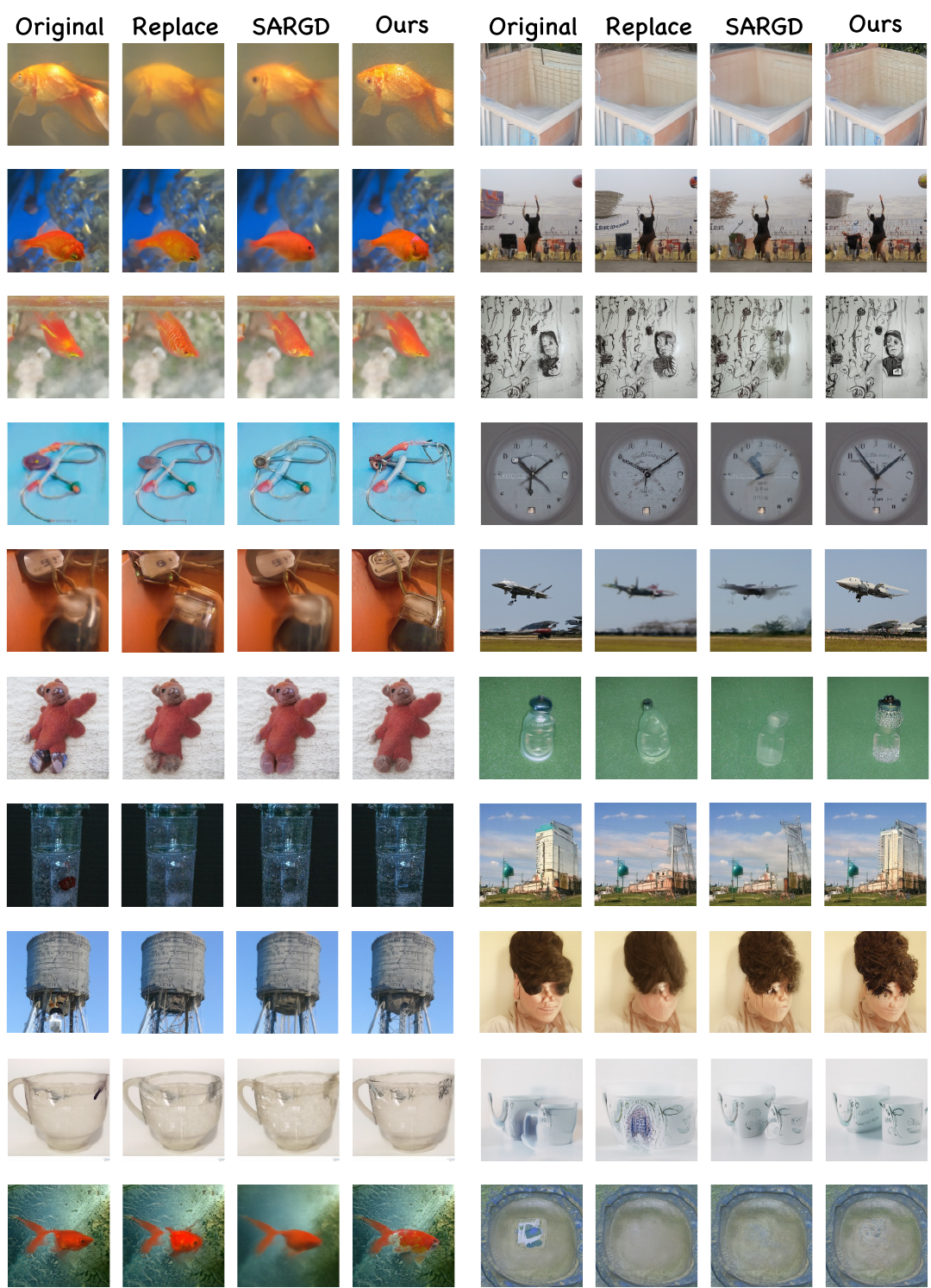}
\vspace{-1 em}
\caption{Additional qualitative comparison of artifact correction methods, following the similar format as \cref{fig:diversity} in the main text.}
\vspace{-1 em}
\label{fig:more_corrected_examples_2}
\end{figure*}

Following the qualitative analysis (\cref{fig:diversity}) in the main paper, we provide additional correction results (\cref{fig:more_corrected_examples_1} and \cref{fig:more_corrected_examples_2}) across different datasets to demonstrate the consistent performance of the proposed method. These examples further illustrate the effectiveness of trajectory-aware target correction (ours) in preserving local details while removing artifacts.

\end{document}